\documentclass[12pt,letterpaper]{article}
\usepackage[margin=1in]{geometry}
\usepackage{setspace}
\usepackage{etoolbox}
\makeatletter
\patchcmd{\@footnotetext}{\def\baselinestretch{\setspace@singlespace}}{\def\baselinestretch{1.655}}{}{}
\patchcmd{\@mpfootnotetext}{\def\baselinestretch{\setspace@singlespace}}{\def\baselinestretch{1.655}}{}{}
\makeatother
\usepackage{color}
\usepackage{multirow}
\usepackage[authoryear,round,sort]{natbib}
\usepackage{rotating}
\usepackage{latexsym}
\usepackage{threeparttable}
\usepackage{xurl}
\usepackage[hidelinks]{hyperref}
\usepackage[expansion=false]{microtype}

\usepackage{graphicx}
\usepackage{tikz}
\usepackage{amsmath, mathtools, xfrac}
\usepackage{txfonts}

\DeclareUnicodeCharacter{2212}{\textminus}
\usepackage{array,booktabs} 

\usepackage[font={normalsize,doublespacing},labelfont=bf,justification=raggedright,singlelinecheck=false]{caption}
\usepackage{wrapfig}
\usepackage{sidecap}
\usepackage{placeins,needspace}

\usepackage{titlesec}
\titleformat{\section}{\normalfont\normalsize\bfseries\sffamily}{\thesection}{1em}{}
\titleformat{\subsection}{\normalfont\normalsize\bfseries\sffamily}{\thesubsection}{1em}{}
\titleformat{\subsubsection}{\normalfont\normalsize\bfseries\sffamily}{\thesubsubsection}{1em}{}

\usepackage{multirow}

\usepackage{fancyhdr}
\usepackage{authblk}

\makeatletter

\renewcommand{\@maketitle}{%
  \begin{center}
    {\normalsize\bfseries\@title\par}
    \vspace{\baselineskip}
    {\normalsize\@author\par}
  \end{center}%
}
\makeatother
\title{ Hierarchical Latent Structure Learning through Online Inference}

\author[1]{Ines Aitsahalia}
\author[1,2,3,4]{Kiyohito Iigaya}
\affil[1]{Center for Theoretical Neuroscience and Zuckerman Institute, Columbia University, New York, NY 10027}
\affil[2]{Department of Psychiatry, Columbia University Irving Medical Center, New York, NY 10032}
\affil[3]{New York State Psychiatric Institute, New York, NY 10032}
\affil[4]{Columbia Data Science Institute, New York, NY 10027}
\date{}
\newsavebox{\manuscriptfigurebox}
\newenvironment{manuscriptfigure}[2]{%
  \par\FloatBarrier
  \sbox{\manuscriptfigurebox}{#2}%
  \Needspace{\dimexpr\ht\manuscriptfigurebox+\dp\manuscriptfigurebox+4\baselineskip\relax}%
  \refstepcounter{figure}\label{#1}%
  \addvspace{\baselineskip}%
  {\centering\noindent\usebox{\manuscriptfigurebox}\par}%
}{\par\addvspace{\baselineskip}}
\newcommand{\manuscriptcaption}[1]{%
  \par\nopagebreak[4]\vspace{0.5\baselineskip}\nopagebreak[4]%
  \noindent\textbf{\figurename~\thefigure:} #1\par
}

\begin{document}

\maketitle
\thispagestyle{fancy}

\section*{Abstract}

Learning systems must balance generalization across experiences with discrimination of task-relevant details. Effective learning therefore requires representations that support both.
Online latent-cause models support incremental inference but assume flat partitions, whereas hierarchical Bayesian models capture multilevel structure but typically require offline inference. We introduce the \textbf{Hierarchical Online Learning of Multiscale Experience Structure (HOLMES) model}, a computational framework for hierarchical latent structure learning through online inference. HOLMES combines a variation on the nested Chinese Restaurant Process prior with sequential Monte Carlo inference to perform tractable trial-by-trial inference over hierarchical latent representations without explicit supervision over the latent structure.  In simulations, HOLMES matched the predictive performance of flat models while learning more compact representations that supported one-shot backward transfer to higher-level latent categories. In a forward transfer task, HOLMES additionally achieved above-chance outcome prediction for stimuli with never-before-seen feature combinations, by exploiting abstract representations learned across diverse training instances. These results provide a tractable computational framework for discovering hierarchical structure in sequential data.

\clearpage
\section*{Introduction}

A central challenge in cognition is balancing \emph{generalization} and \emph{discrimination}. Generalization allows agents to apply prior knowledge to novel situations, whereas discrimination enables sensitivity to subtle but meaningful differences in observations \citep{tenenbaum_generalization_2001,tenenbaum2011grow,Barak2013,Lake2015}. Effective learning therefore requires representations that capture shared structure across multiple levels of abstraction while preserving task-relevant distinctions among observations.

One influential approach is latent structure inference, in which an agent partitions observations into hidden causes that generate the data. Bayesian nonparametric models provide a principled framework for this process by allowing the number of latent causes to grow with experience \citep{anderson1991adaptive,mansinghka2006structured,gershman_context_2010,gershman_exploring_2012}.  
However, most existing models assume flat partitions in which all latent causes exist at a single level of abstraction.

Many environments are instead inherently \emph{hierarchical}~\citep{ODoherty2021},
where observations share intermediate categories and broader contextual regularities.
Hierarchical Bayesian models \citep{Teh2006}
such as the nested Chinese Restaurant Process (nCRP) \citep{griffiths2003hierarchical,blei2010nested} capture such structure,
but typically rely on offline batch inference.
Conversely, online latent cause models based on sequential Monte Carlo methods support cognitively relevant, incremental trial-by-trial inference but typically assume flat latent spaces~\citep{gershman_tutorial_2012}.
Bridging this gap requires a model
that combines hierarchical representation with online inference.

Here, we introduce \emph{Hierarchical Online Learning of Multiscale Experience Structure (HOLMES)}, a framework that combines hierarchical nonparametric structure with online sequential inference.
Our model performs trial-by-trial inference over dynamically expanding latent trees, enabling hierarchical representations to be discovered directly from sequential experience.  

We evaluate the model in two classes of synthetic tasks. In compositional environments, hierarchical inference preserves predictive performance while learning more compact representations that support one-shot transfer across latent categories. In a novel-exemplar generalization task, hierarchical inference additionally enables above-chance outcome prediction for previously unseen feature
combinations by using higher-level structure to bias new stimuli toward
previously learned, outcome-informative leaves.

\section*{Methods}
 
\subsection*{Formal problem definition}
We consider a task in which an agent observes a sequential dataset
\begin{equation}
D = \{d_1, d_2, \dots, d_N\},
\end{equation}
where each observation \( d_i \) is an \( F \)-dimensional binary vector indicating the presence (1) or absence (0) of each feature at time \( i \). One feature dimension, denoted $o$, is designated the \emph{outcome}; the remaining dimensions are \emph{cues}. In classical conditioning, for example, the cues may encode the presence of a conditioned stimulus, and the associated outcome.

The agent's goal is not simply to partition observations into latent causes, but to use that partition to predict the outcome dimension of each new observation before it is revealed. This requires inferring, at each time step $i$, a latent assignment sequence
\begin{equation}
\mathbf{Z} = \{Z_1, Z_2, \dots, Z_N\},
\end{equation}
where $Z_i \in \{1, 2, \dots, K\}$ indexes the cluster associated with observation $i$. Critically, both the number of clusters $K$ and their assignments are unknown and must be inferred from data. Cluster assignment enables this prediction since each cluster maintains a running estimate of the distribution over all $F$ features, and it is this estimate that yields the agent's trial-by-trial predictions about the features of the upcoming observations. 

Because the number of possible partitions grows super-exponentially, exact inference is intractable \citep{Hjort_Holmes_Müller_Walker_2010}. We therefore combine a Bayesian nonparametric prior \citep{Aldous1985} and a sequential Monte Carlo method (particle filtering) \citep{speekenbrink_tutorial_2016, Martino2018-st} to approximate posterior beliefs online.
 
Prediction and learning proceed in two stages on each trial. First, before the outcome is revealed, the agent forms a prediction $\hat{r}_i$ as a posterior expectation over the outcome feature $o$, using sufficient statistics accumulated from all prior trials and weighting particles by the likelihood of the \emph{non-outcome} features only:

%changed to separate full-outcome from partial outcome

\begin{align}
    \tilde{w}^{(p)}_i &= \frac{L_{\mathcal{F}\setminus\{o\}}\!\big(\mathbf{d}_i \mid Z^{(p)}_i\big)}
                              {\sum_{p'} L_{\mathcal{F}\setminus\{o\}}\!\big(\mathbf{d}_i \mid Z^{(p')}_i\big)},
    \label{eq:predweight}\\[10pt]
    \hat{r}_i &= \sum_p \tilde{w}^{(p)}_i \cdot
        \frac{n^{(i-1)}_{o,\, Z^{(p)}_i}}{n^{(i-1)}_{o,\, Z^{(p)}_i} + b^{(i-1)}_{o,\, Z^{(p)}_i}},
    \label{eq:prediction}
\end{align}
where $\mathcal{F} = \{1,\dots,F\}$ is the full feature set, $o$ is the outcome feature, and $L_{\mathcal{F}\setminus\{o\}}$ is the factorized feature likelihood evaluated over all features except $o$ (Eq.~\ref{eq:likelihood}). The outcome probability is read off the cluster's current sufficient statistics \emph{before} the trial's outcome is incorporated, so $\hat{r}_i$ is a genuine one-step-ahead prediction. Second, once the outcome is observed, the agent updates the sufficient statistics for \emph{all} features --- including the outcome --- for the cluster assigned to trial $i$:
\begin{align}
    n^{(i)}_{f,k} &= n^{(i-1)}_{f,k} + d_{i,f}, \quad \forall\, f \in \{1, \dots, F\}, \label{eq:nupdate_full} \\
    b^{(i)}_{f,k} &= b^{(i-1)}_{f,k} + (1 - d_{i,f}), \quad \forall\, f \in \{1, \dots, F\}. \label{eq:bupdate_full}
\end{align}

\subsection*{Flat latent-cause model}
 
In the flat latent-cause model, observations are grouped using a Bayesian nonparametric clustering formalism, with clusters commonly interpreted as ``latent causes'' that generate the observations \citep{gershman_context_2010}. On each trial, the model evaluates whether the current observation is better explained by an existing cluster or by introducing a new one.
 
\paragraph{Prior over cluster assignments}
 
The prior over latent assignments is specified using the Chinese Restaurant Process (CRP) (Fig.~\ref{fig:crp}A). This allows the number of clusters (tables in a restaurant) to grow with experience while favoring reuse of previously inferred clusters \citep{Aldous1985}.

\begin{figure}
    \centering
    \includegraphics[width=1\linewidth,height=0.54\textheight,keepaspectratio]{NECO-D-26-00058R2-Figure.1.eps}
    \caption{Online hierarchical extension of the Chinese Restaurant Process (CRP) for structure learning.
(A) CRP Algorithm. The CRP provides a Bayesian nonparametric prior over partitions, assigning each observation i (filled circle) to an existing cluster k  with probability proportional to its occupancy ($n_k$) and concentration parameter $\alpha$ or to a new cluster.
(B) HOLMES prior with nested Chinese Restaurant Process (nCRP) algorithm. The nCRP provides a Bayesian nonparametric prior over tree-structured partitions, assigning observations to paths by sequentially selecting branches at each level $L$ with probability proportional to occupancy and  $\alpha_L$, the level-specific concentration.}
    \label{fig:crp}
\end{figure}

Under the CRP, the probability that observation $i$ is assigned to an existing cluster $k$ or a new cluster is:
\begin{equation}
P(Z_i = k \mid Z_{1:i-1}) = \frac{n_k}{i - 1 + \alpha}, \quad
P(Z_i = \text{new} \mid Z_{1:i-1}) = \frac{\alpha}{i - 1 + \alpha},
\end{equation}
where $n_k$ is the number of previous observations assigned to cause $k$, and $\alpha$ is a concentration parameter. Larger $\alpha$ favors finer partitions with more clusters, whereas smaller $\alpha$ favors reuse of existing clusters and broader generalization.

\paragraph{Likelihood computation and online updates}

Given current feature counts, the per-feature Beta--Bernoulli predictive probability of observation $\mathbf{d}_i$ under cluster $k$ is
\begin{equation}
p_{f,k}^{(i)} =
\begin{cases}
\dfrac{n_{f,k}^{(i-1)}}{n_{f,k}^{(i-1)} + b_{f,k}^{(i-1)}} & \text{if } d_{i,f} = 1, \\[20pt]
\dfrac{b_{f,k}^{(i-1)}}{n_{f,k}^{(i-1)} + b_{f,k}^{(i-1)}} & \text{if } d_{i,f} = 0,
\end{cases}
\label{eq:bernoulli_pred}
\end{equation}

All feature counts are initialized with a symmetric pseudocount $\Omega > 0$:
\begin{equation}
n_{f,k}^{(0)} = \Omega, \quad b_{f,k}^{(0)} = \Omega. \label{eq:init_nb}
\end{equation}
Assuming conditional independence of features given cluster assignment, the joint likelihood over a feature subset $\mathcal{F}' \subseteq \mathcal{F}$ is
\begin{equation}
L_{\mathcal{F}'}\!\big(\mathbf{d}_i \mid Z_i = k\big) = \prod_{f \in \mathcal{F}'} p_{f,k}^{(i)},
\label{eq:likelihood}
\end{equation}
where $p_{f,k}^{(i)}$ is given by Eq.~\ref{eq:bernoulli_pred}, and counts are initialized with the pseudocount $\Omega$ (Eq.~\ref{eq:init_nb}) and include all observations up to observation $i-1$.
 
This pseudocount parameter $\Omega$ controls the influence of new evidence: larger $\Omega$ promotes slower updating and greater generalization, whereas smaller $\Omega$ allows more rapid context differentiation at the cost of increased sensitivity to noise and risk of overfitting.

Since particles compete through their relative likelihoods, the raw product makes the weights scale with the number of features evaluated so we normalize these when the task has redundant features.
 
\paragraph{Sequential Monte Carlo inference}
 
Exact posterior inference over latent assignments is intractable. We therefore approximate the posterior using particle filtering. The particle filter maintains an ensemble of $P$ particles, each representing a hypothesis about the current clustering state, including cluster assignments and feature statistics.
 
At observation $i$, particles sample assignments from the CRP prior and are weighted by the likelihood of the current observation:

% update with the new notation
\begin{equation}
w_i^{(p)} \;\propto\; L_{\mathcal{F}_w}\!\big(\mathbf{d}_i \mid Z_i^{(p)}\big),
\qquad
\mathcal{F}_w =
\begin{cases}
\mathcal{F} & \text{on feedback trials},\\
\mathcal{F}\setminus\{o\} & \text{otherwise},
\end{cases}
\label{eq:particle_weight}
\end{equation}
with weights normalized across particles. Particles are resampled at every observation to counter degeneracy.

The model's predictions (Eq.~\ref{eq:prediction}) are posterior expectations over the particle ensemble, naturally reflecting uncertainty when particles disagree.
 
\subsection*{HOLMES: online hierarchical latent structure learning}
 
We extend the flat latent-cause model by assigning observations to paths through a hierarchical tree of latent causes. Unlike standard flat formulations, this extension organizes latent structure across multiple levels of abstraction, allowing higher-level nodes to capture shared regularities and lower-level nodes to capture more detailed distinctions.

\paragraph{Hierarchical prior over latent paths}
 
We define the prior over tree structures using a \emph{depth-decayed, sticky hierarchical Chinese Restaurant Process}, inspired by the nested CRP \citep{blei2010nested,griffiths2003hierarchical} but with branch-selection statistics shared across nodes at the same depth. Unlike a standard CRP, in which observations are assigned to a single flat partition, our prior assigns each observation to a \emph{path} through a latent tree (Fig.~\ref{fig:crp}B). Throughout, we use \emph{hierarchy} to mean a rooted tree in which representations become progressively more specific as one descends from the top of the tree. A \emph{node} is a latent cause at a particular level, and a \emph{branch} is an outgoing choice from one node to a child node at the next level. Each observation is assigned to a \emph{path}, defined as the sequence of nodes visited from the top of the tree downward. The \emph{leaf} is the terminal node of that path---that is, the node at which traversal stops for the current observation. Because stopping is stochastic, leaves can occur at different depths for different observations. In the figures, we therefore draw broader shared structure toward the top and increasingly specific distinctions toward the bottom.

Like other Bayesian latent-structure models, HOLMES assumes a specified family of observation models and latent representations. The structural quantities treated as unknown are the assignments of observations within the hierarchy and the degree of hierarchical elaboration: both the occupied branching structure and the effective depth of the representation are inferred online from sequential experience.

At each trial $t$, each particle $p$ samples a sequence of latent assignments
\[
\mathbf{c}_t^{(p)} = \big(c_t^{(p,0)}, c_t^{(p,1)}, \dots, c_t^{(p,L-1)}\big),
\]
where $c_t^{(p,\ell)}$ indexes a branch selected at level $\ell$, and the depth $L$ may vary across trials.
 
\paragraph{Branch selection.}
At level $\ell$, the probability of assigning an observation to an existing branch index $k$ or creating a new one follows the CRP form with depth-adjusted concentration:
\begin{align}
P(c_t^{(p,\ell)} = k) &= \frac{n_{k}^{(p,\ell)}}{N^{(p,\ell)} + \alpha_\ell},
\quad
P(c_t^{(p,\ell)} = \text{new}) = \frac{\alpha_\ell}{N^{(p,\ell)} + \alpha_\ell},
\label{eq:ncrp_existingnew}
\end{align}
where $n_{k}^{(p,\ell)}$ is the number of previous assignments to branch index $k$ at level $\ell$ within particle $p$, pooled across all parent nodes at that depth, and $N^{(p,\ell)} = \sum_k n_{k}^{(p,\ell)}$. The node identity visited by following branch $k$ from parent $u$ at level $\ell$ is determined by the $(\ell, u, k)$ triple via a canonical node registry shared across particles (Supplementary Methods); the same branch index under different parents therefore instantiates different latent causes. This level-sharing of sampling statistics introduces an inductive bias toward parsimonious trees: branches discovered under one parent inform branch selection under others, encouraging reuse of a small canonical set of branch indices at each depth.

Suppose that at level $\ell$, branch index $k=0$ has been selected frequently under several different parent nodes, whereas branch index $k=1$ has rarely been selected. Because the counts $n_k^{(p,\ell)}$ are pooled across parents at that depth, a particle arriving at another parent at level $\ell$ will have a prior preference for selecting branch index $0$. Importantly, the branch index is a reusable \emph{branch label}, not a globally shared child node: because node identity is defined by the full $(\ell,u,k)$ triple, choosing $k=0$ beneath two different parents produces two distinct latent nodes. What is shared across parents is therefore the prior over branch choices, not the observations or sufficient statistics stored at those nodes. This induces a parsimony bias toward reusing a small set of branching patterns at a given depth while preserving parent-specific latent causes.

We cap the number of distinct branch indices at each level and the maximum depth as 20, as a weak limit approximation despite possible unbounded growth. The first branch instantiated at any level is created deterministically (the degenerate case $\alpha_\ell/(0+\alpha_\ell)=1$).
 
\paragraph{Depth-decayed concentration and stopping.}
 
We define depth-adjusted concentration as
\begin{equation}
\alpha_\ell = \alpha \cdot e^{-\alpha \cdot \ell},
\label{eq:depth_decay}
\end{equation}

where the exponential decay rate is tied to the base concentration ($\texttt{depth\_decay}=\alpha$ in the implementation). This implements a \textit{depth-budget constraint}: models with higher $\alpha$ (favoring proliferation of clusters) deplete their capacity for deep branching more rapidly, so the model cannot simultaneously maintain high branching factors and deep hierarchies---a form of implicit capacity control. Under this parameterization, $\alpha_\ell$ is non-monotone in $\alpha$ for fixed $\ell > 0$ (maximized at $\alpha = 1/\ell$), reflecting the tradeoff between local branching propensity and global depth budget. We interpret $\alpha$ as controlling overall model capacity rather than concentration at any single level.
 
In addition to depth-dependent branching, traversal can terminate stochastically at \textbf{any depth} \citep{ghahramani2010tree}. Crucially, this allows the model to adaptively pick depth, improving inference performance by better matching the generative task structure (Fig.~\ref{fig:ablation}). Level $\ell = 0$ is always instantiated (no stopping test is applied there), ensuring that at least one representational level exists. For $\ell \ge 1$, having instantiated level $\ell-1$, traversal terminates (returning a leaf at level $\ell-1$) with probability
\begin{equation}
P(\text{stop before level } \ell \mid \text{reached level } \ell-1)
   = \frac{1}{1 + \alpha_{\ell-1}},
   \qquad \ell \ge 1,
\label{eq:stopping}
\end{equation}
and otherwise descends to level $\ell$ and selects a branch via Eq.~\ref{eq:ncrp_existingnew}. The concentration governing the stop test is that of the just-instantiated parent level, $\alpha_{\ell-1} = \alpha\,e^{-\alpha(\ell-1)}$; because $\alpha_\ell \to 0$ with depth, the stopping probability approaches $1$ and paths terminate without an explicit depth limit. 

\paragraph{Persistence and node reuse}

To capture temporal persistence in latent structure, we include a stickiness bias \citep{Fox2011} tied to the feature pseudocount $\Omega$ that increases the probability of reusing the previously inferred latent configuration at each level and allows reuse of global node identities across particles. When different particles discover identical substructures, they share the same global node rather than maintaining redundant local copies \citep{blei2010nested} (see Supplementary Methods for full implementation and necessity, Fig.~\ref{fig:ablation}). 

\paragraph{Hierarchical inference procedure}
Inference used the same particle-filtering framework as the flat model. At each trial, each particle samples a path through the hierarchical prior, evaluates the likelihood of the observation at the \emph{leaf} node of that path (Eqs.~\ref{eq:bernoulli_pred}--\ref{eq:likelihood}), and updates its weight (Eq.~\ref{eq:particle_weight}) accordingly; weights are normalized and particles resampled to approximate the posterior. The prediction $\hat{r}_i$ uses the cue-only weights of Eq.~\ref{eq:predweight} read at the leaf, exactly as in the flat model.
 
% clarified the leaf node
Although likelihoods are evaluated only at the leaf nodes, and internal nodes maintain no feature or outcome sufficient statistics, the upper levels shape inference entirely through the prior. Because branch-selection counts are pooled across parents at each depth and node identities are shared across particles, observations in the same high-level branch have a higher probability of being assigned to the same descendant leaf, both through this prior and the stickiness (Eq.~\ref{eq:stickiness}). Transfer across surface variation therefore arises from prior-driven reuse of an existing leaf---which already carries that lineage's accumulated outcome statistics---rather than from evidence pooling at internal nodes. This keeps the inference space constrained even as the hypothesis space grows.
 
\subsection*{Simulation environment}

\paragraph{Compositional task environment}
We first evaluated the models in synthetic compositional tasks (Fig.~\ref{fig:compression}A). These tasks contained two to five levels, where each level represented a binary latent category that determined observable feature values. Observations were binary feature vectors generated by level-specific rules. The top $n-2$ levels were encoded as single binary features, the next lower identity was encoded with two binary features, and the lower levels were encoded as a one-hot vector across four random features. Observations were generated by sampling a latent category value at each hierarchical level, then deterministically encoding the resulting configuration as features. Outcomes were determined by a conjunctive rule over the latent category assignments at the two highest levels (Fig.~\ref{fig:compression}A).

On each trial, the agent receives the full stimulus vector, including the outcome dimension, and predicts the outcome from the remaining (cue) features before that outcome is revealed, as described above (Eqs.~\ref{eq:predweight}--\ref{eq:prediction}); the true outcome is then incorporated to update the model's beliefs (Eqs.~\ref{eq:nupdate_full}--\ref{eq:bupdate_full}). Feedback is available on every trial in this task.

\paragraph{Novel color generalization task}
We next evaluated forward transfer to unseen stimuli (Fig.~\ref{fig:rule}A). Reward depended on a single relevant dimension (shape) while a second dimension (color) varied more saliently. The environment contained two shapes and 10 colors. 8 of these appeared during training and the remaining 2 were held out for test. Each stimulus was also a binary feature vector in which shape and color were encoded by one-hot copies, followed by a single outcome feature. Reward was deterministic, appearing if and only if the shape matched a designated rewarded shape, independent of color. This was to test rule learning.
 
During training, each pairing of the two shapes with each of the eight
training colors was presented 10 times, yielding 160 training trials.
At test, each pairing of a shape with a held-out color was presented three
times. Analyses of zero-shot transfer used only the first presentation of
each held-out shape--color pairing. The primary forward transfer metric was the model's outcome prediction on the \emph{first} presentation of each held-out color, before any feedback had been received for that color, isolating zero-shot generalization of the shape$\rightarrow$reward rule to a novel stimulus.

\subsection*{Evaluation metrics}

\paragraph{Outcome prediction accuracy}
Outcome prediction accuracy was the proportion of trials on which the model's predicted reward probability fell on the correct side of $0.5$ relative to the realized outcome.

\paragraph{Representational efficiency}
To quantify representational efficiency, we measured the entropy of cluster assignments within each ground-truth category: $H_m = -\sum_k p_{mk} \log p_{mk}$, where $p_{mk}$ is the proportion of category-$m$ trials assigned to cluster $k$. Lower entropy indicates that trials from a category concentrated in fewer clusters. In the transfer task, categories corresponded to ground-truth labels at a given hierarchical level (see Supplemental for details). 

\paragraph{Backwards transfer}
To assess generalization, we evaluated one-shot transfer after training. A single higher-level labeled exemplar was presented and models were required to generalize that label to previously observed stimuli belonging to the same latent category. Transfer performance was quantified using recall: the proportion of true same-category trials correctly identified \citep{precision}. We focus on recall as the primary metric because (1) category membership is balanced across trials, making recall directly interpretable as generalization success, and (2) the scientific question concerns the model's ability to recognize same-category instances across contexts, which recall measures directly. This is highly correlated with F1 score across all task complexities (for additional metrics, see Supplementary Table~2).

\paragraph{Forward transfer to novel stimuli}
To assess whether a learned shape to reward association generalized to entirely unseen stimuli, we evaluated held-out trials in which a familiar shape appeared in a novel color. On each trial the model produced a graded reward prediction, and we scored discrimination between rewarded-shape and other-shape stimuli using the area under the receiver operating characteristic curve (AUC). AUC is the probability that a randomly chosen rewarded-shape trial receives a higher predicted reward than a randomly chosen other-shape trial; it therefore quantifies how well the model \emph{ranks} novel stimuli by reward rather than whether its predictions cross a particular decision threshold. We report AUC alongside thresholded accuracy because the two answer complementary questions: accuracy measures whether each prediction falls on the correct side of the
0.5 boundary (the Bayes-optimal threshold given balanced reward), whereas AUC integrates over all possible thresholds and is therefore invariant to the absolute scale of the predictions. 

\paragraph{Shape separation metrics}
To characterize representational structure, we computed a \emph{shape separation} statistic from each model's predicted reward probabilities. For every first-exposure held-out trial, we centered the predicted probability of reward on the decision boundary, expressing each prediction as a signed distance from indifference in percentage points.  Trials were then grouped by ground truth into those belonging to the rewarded shape and those belonging to the other shape, and shape separation was defined as the difference in mean centered score between these two groups, $\Delta = \bar s_{\text{rewarded}} - \bar s_{\text{other}}$. A model that reliably assigns higher reward probability to the rewarded shape than to the other shape, regardless of where exactly its predictions fall relative to 0.5, will show a large positive $\Delta$; a model with no shape-specific signal will show $\Delta \approx 0$.

\section*{Results}

\subsection*{Hierarchical inference preserves outcome prediction while improving representational efficiency}

We first examined whether hierarchical inference improves the efficiency with which compositional structure is represented during learning. In many environments, observable features combine to form higher-level latent categories that determine outcomes. Learning in such settings requires discovering how observations compose into reusable abstractions. We therefore asked whether hierarchical clustering produces more efficient internal representations than flat clustering when learning compositional tasks.

To test this, we constructed synthetic categorization environments in which observations were generated from latent factors organized in a nested hierarchy (Fig. ~\ref{fig:compression}A).  In the simplest two-level task, each of four observations, represented as a one-hot vector, belonged to one of two latent categories, such as 'A' or 'B'. Reward depended only on category: stimuli from group A were rewarded, whereas stimuli from group B were not, regardless of the particular exemplar presented. 

To make the learning problem concrete, consider the 3-level example in Fig.~\ref{fig:compression}A. On each trial the model sees only one colored shape and must predict its outcome before reward is revealed. For example, it may first encounter a red rectangle with reward; after making its prediction, the model updates the leaf assigned to that observation with the observed features and outcome. On a later trial it may encounter a pink rectangle, again followed by reward, and subsequently a magenta or orange triangle without reward. The flat model must explain each complete feature vector using a single level of abstraction, which is easy via simple memorization task or binary classification (squares vs. all other stimuli). HOLMES instead considers variable-depth paths through the tree, allowing a particle to represent shared regularities at one level while retaining stimulus-specific distinctions farther down the path. Across repeated observations, posterior weighting and resampling favor representations that efficiently account for the recurring feature--outcome structure. Crucially, labels such as "rectangle", "triangle", or the higher-order groups depicted in the figure are never supplied to either model during this learning phase; they are visual descriptions of structure that may emerge from the observations.

Despite their architectural differences, both models achieved equivalent performance on the primary learning objective. We quantified learning performance as outcome prediction accuracy—the proportion of trials on which models correctly predicted binary outcomes (reward vs. no reward) from observable features. At 3-level complexity, flat and hierarchical models showed statistically indistinguishable accuracy (Fig. ~\ref{fig:compression}B). Thus hierarchical organization does not impair predictive learning.

However, the models differed markedly in representational efficiency. To quantify, we measured the entropy of cluster assignments within each true category: a metric capturing how consistently trials from the same category mapped to the same cluster. Lower entropy indicates more organized, efficient representations. HOLMES showed significantly lower entropy than flat models at the 3-level task (Fig. ~\ref{fig:compression}C), indicating more consistent category-to-cluster mappings despite equivalent outcome prediction.

\begin{manuscriptfigure}{fig:compression}{%
    \includegraphics[width=1\linewidth,height=0.67\textheight,keepaspectratio]{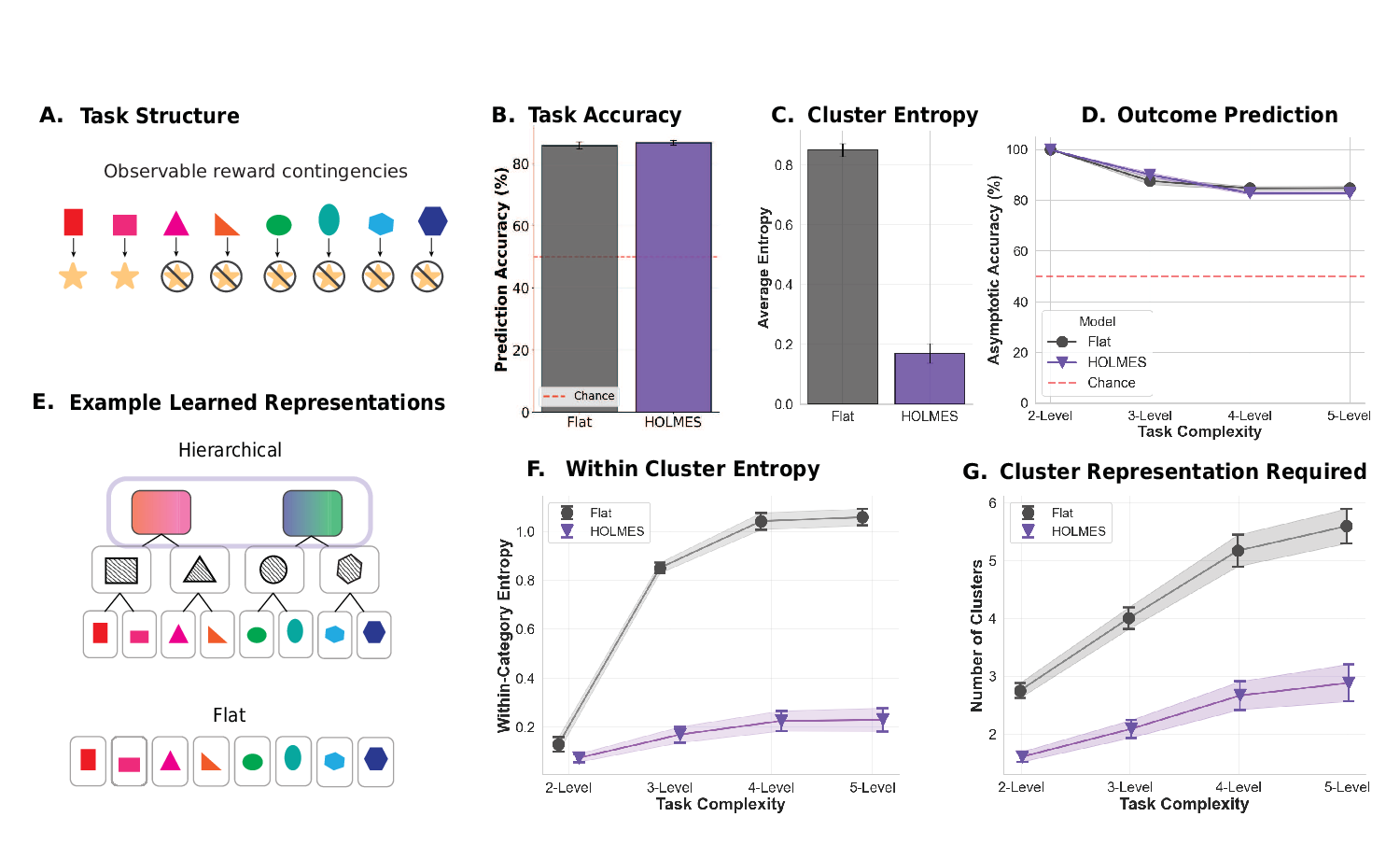}}
    \manuscriptcaption{Hierarchical inference preserves outcome prediction accuracy while improving representational efficiency. (A)  Reward prediction task structure. Schematic depiction of the 3-level task. 8 separate stimuli are shown with some overlapping features (here, shape and color). Only 2 stimuli, the red and pink rectangles, are paired with reward (stars). On any given trial, the model must predict whether or not a reward will appear based on the single stimulus shown. The number of stimuli presented scales with the number of levels in the generative structure, with the simplest flat task only having 4.  
    (B) Outcome prediction performance on 3-level task. Both models achieve equivalent high accuracy. Error bars: 95\% CI across 200 parameter combinations.
    (C) Representational efficiency measured by average entropy of cluster assignments. HOLMES (hierarchical models) shows significantly lower entropy (0.076±0.017 vs. 0.131±0.030), indicating better compression. Error bars: 95\% CI across 200 parameter combinations.
    (D) Outcome prediction performance across task complexity. Both models achieve comparable asymptotic accuracy (84-100\%) across all complexity levels, with small effect sizes (Cohen's $|d|$ < 0.4). Error bars: 95\% CI across 200 parameter combinations.
    (E) Example learned model structures for the 3 level task. The flat latent-cause model can represent each observation type separately at the observation level, while HOLMES learns a compressed category-level representation. Here, each stimulus can belong to both a shape-level cluster and a color-level cluster.
    (F) Representational efficiency across complexity. HOLMES shows significantly lower entropy at all levels, with increasing advantage at higher complexities (2-level: -0.055; 5-level: -0.827). Error bars: 95\% CI across 200 parameter combinations.
    (G) Number of learned clusters across task complexity. HOLMES (light) maintains near-optimal compression across complexities, while flat models (dark) show increasing redundancy, using significantly more clusters at all levels (differences: -1.1 to -2.7 clusters). Error bars: 95\% CI across 200 parameter combinations.}
\end{manuscriptfigure}

We next examined how these effects scale with task complexity.  We generated tasks with up to 5 hierarchical levels, training both models to predict the same binary outcome across increasing structural complexity. Across all complexity levels, both models achieved comparable accuracy in predicting outcomes  (Fig. ~\ref{fig:compression}D). However, HOLMES showed significantly lower entropy than flat models at all complexity levels (Fig. ~\ref{fig:compression}F), indicating more consistent category-to-cluster mappings regardless of task depth.

We further quantified compression efficiency by measuring the number of clusters learned relative to the true number of latent categories. HOLMES maintained near-optimal compression across task complexities, using approximately as many clusters as true category labels (Fig.~\ref{fig:compression}G, light). In contrast, flat models showed increasing redundancy with complexity, requiring progressively more clusters than necessary (Fig.~\ref{fig:compression}G, dark). Together, these results demonstrate that hierarchical clustering preserves predictive performance while learning substantially more efficient internal representations of compositional structure.

To ensure this was not merely a result of task structure, we evaluated both models on a task with a flat generative structure, showing that HOLMES' adaptive depth also allows for equivalent accuracy and representational efficiency on flat tasks (Fig.~\ref{fig:cc}A). 

\subsection*{Hierarchical representations enable one-shot transfer across latent categories}

\begin{manuscriptfigure}{fig:generalization}{%
    \includegraphics[width=0.8\linewidth,height=0.67\textheight,keepaspectratio]{NECO-D-26-00058R2-Figure.3.eps}}
    \manuscriptcaption{Hierarchical advantage in one-shot transfer emerges with task complexity.
(A) Backward-transfer task structure. The schematic represents the 3-level task. During the learning phase, the model observes colored shapes and their outcomes and uses this experience to construct a latent representation of the task; no category labels are provided during learning. After learning, an exemplar--for example, the red rectangle-- is labeled as belonging to a group, such as one with the magenta triangle. This label is then used only as a probe: the model is asked which other previously observed exemplars belong to that same group. In the example shown, despite only the triangles leading to reward, HOLMES can learn a hierarchical representation of ‘color,’ with the red and pink rectangles together with the magenta and orange triangles under a common higher-order branch, and therefore predicts that the pink rectangle and orange triangle also belong to the labeled group. The label does not create this grouping; it tests whether the grouping was already present in the structure learned from experience. Task difficulty scales with the tree complexity. 
(B) One-shot transfer across task complexity. While both models showed a decrease in recall accuracy as task complexity increased, HOLMES (light, triangle) maintained superior transfer at higher complexities, while flat models (dark, circle) showed declining performance. At 2-level complexity, both models performed comparably (flat: 89.7±2.0\%, hierarchical: 89.3±1.8\%); however, hierarchical models significantly outperformed 
flat models at higher complexities (3-level: +21.0\%, 95\% CI [19.2\%, 22.8\%]; 4-level: +24.7\%, 95\% CI [22.6\%, 26.8\%]; 5-level: +26.6\%, 95\% CI [24.4\%, 28.8\%]. Error bars represent 95\% confidence intervals across 200 parameter combinations.
(C) Hierarchical advantage (hierarchical minus flat transfer accuracy) 
increases systematically with task complexity, transitioning from no advantage at 2-level to substantial advantages at 3+ levels.
Error bars represent 95\% confidence intervals across n=200 parameter combinations.
(D) One-shot transfer performance at the most complex task for all the levels tested. The HOLMES (hierarchical model; right) outperforms the flat model, with greater advantage on deeper levels. 
(E) Relationship between outcome prediction accuracy and one-shot transfer 
accuracy across 200 parameter settings for hierarchical (purple) and flat models (gray). Each point represents one parameter combination averaged over seeds. Regression lines show 95\% confidence bands.
}
\end{manuscriptfigure}

If hierarchical inference discovers latent compositional structure during learning, the resulting representations should support rapid generalization across observations sharing the same latent category. We therefore next asked whether the hierarchical model's more efficient representations enable one-shot transfer across latent categories.

To test this prediction, we implemented a one-shot transfer paradigm. Models were first trained on outcome prediction exactly as before, learning to predict reward outcomes without ever receiving explicit category labels. After training, we introduced a teaching phase consisting of a single labeled example—showing the model one previously-observed stimulus and labeling it as belonging to a particular latent category. In the transfer test, models had to generalize this label to identify which other previously-observed stimuli belonged to the same category, despite having only seen a single labeled example and never being trained on categorization directly.
For example, in Fig.~\ref{fig:generalization}A, suppose the labeled exemplar is the red rectangle and the probe concerns the higher-order group illustrated in the learned hierarchy. If the model has already organized the red and pink rectangles together with the magenta and orange triangles under the same higher-order branch, the single label attached to the red rectangle transfers immediately to the other members of that branch. A flat representation can only support the same inference when the relevant category coincides with one of its learned partitions, but it cannot simultaneously expose multiple nested groupings through a single flat partition. For the flat model, a stimulus is classified as belonging to the same category when it occupies the same learned cluster as the labeled exemplar. For HOLMES, the analogous comparison is made at the hierarchical level being probed: stimuli assigned to the same node as the labeled exemplar at that level are classified as members of the same category.

The transfer probe does not alter or retrain the representation learned during outcome prediction. Rather, it asks whether a category relation is already encoded in that representation. In the implementation, one previously observed exemplar from the target category is labeled after learning. This paradigm requires models to have spontaneously discovered the latent category structure during outcome prediction learning. As task depth increases, the transfer probe can therefore ask about progressively broader levels of the learned structure. A model that only learned observation-level associations must treat each stimulus independently and therefore cannot reliably generalize categories. In contrast, a model that learned abstract category-level representations should readily transfer the label across different colored instances of the same latent structure.

Consistent with this prediction, HOLMES showed stronger transfer performance than the flat model as task complexity increased (Fig. ~\ref{fig:generalization}B). While both models performed similarly in the simplest 2-level task, HOLMES substantially outperformed the flat model as the tasks became deeper (Fig. ~\ref{fig:generalization}B). The hierarchical advantage increased systematically with task complexity (Fig.~\ref{fig:generalization}C), indicating that the benefit of hierarchical representations becomes more pronounced as latent structure deepens.

To examine generalization across different abstraction levels, we tested both models on all hierarchical levels in the most complex (5-level) task. The hierarchical model outperformed the flat model across all tested levels, with the advantage being most pronounced at deeper levels of the hierarchy (Fig. ~\ref{fig:generalization}D).

We next examined the relationship between outcome prediction accuracy and one-shot transfer performance across parameter regimes. Both model architectures exhibited a negative correlation between these objectives, suggesting a fundamental tradeoff between fine-grained discrimination and broad generalization. Parameter regimes that optimize outcome prediction create many fine-grained observation-level clusters, which can impair generalization. However, at matched levels of outcome prediction accuracy, hierarchical models consistently achieved superior transfer performance, suggesting a Pareto improvement rather than a simple parameter-dependent tradeoff (Fig.  ~\ref{fig:generalization}E). This advantage likely reflects the hierarchical model's ability to simultaneously maintain task-relevant fine-grained distinctions at lower tree levels while building reusable abstractions at higher levels—a form of representational specialization unavailable to flat architectures.

To assess whether these advantages depended on specific parameter choices, we evaluated model performance across a broad range of concentration and bias parameters. Across 200 sampled parameter combinations, HOLMES consistently produced more efficient representations (Supplementary Fig.~\ref{fig:sup_adv} A,B) and higher transfer accuracy than the flat latent-cause model (Supplementary Fig.~\ref{fig:sup_adv} C,D). These results indicate that the hierarchical advantage reflects architectural properties of the model rather than parameter tuning, and were not rescued by adding stickiness to the flat model (Supplementary Fig.~\ref{fig:ablation}A).

Together, these findings show that hierarchical structure learning enables agents to organize experience into reusable abstractions that support both accurate prediction and rapid transfer from minimal supervision.

\subsection*{Hierarchical inference enables forward generalization}

The preceding sections established that HOLMES forms more compressed latent
representations and, when given a single labeled exemplar, transfers that label
back to other members of the same category. Here we ask: does
the hierarchy support forward transfer?

We tested this in the shape--color task (Fig.~\ref{fig:rule}A), in which reward depended only on shape and color was irrelevant. Models were trained on shape--color combinations for eight
colors and probed on two held-out colors. Accuracy was measured as the predicted
reward on the \emph{first} presentation of each held-out color, before any feedback
for that color had been received. The flat model performed near chance (Fig.~\ref{fig:rule}B), whereas HOLMES generalized highly accurately (Fig.~\ref{fig:rule}B). 

HOLMES achieved near-perfect discrimination between rewarded- and unrewarded-shape trials across all thresholds (AUC $=0.985$), while the flat model's AUC of $0.583$, though modestly above chance, indicates that its probability estimates carry only weak shape-related ranking signal (Fig.~\ref{fig:rule}C).

The source of this difference is illustrated in possible learned cluster structures (Fig.~\ref{fig:rule}D). Because the flat model represents each observation as a single undifferentiated cluster over the full feature vector, it tends to form clusters around specific shape--color conjunctions encountered during training; when a novel color appears, its color features contribute unfamiliar likelihood, so the assignment defaults toward the model's baseline prior rather than an established, reward-informative cluster. This effect was worsened with stickiness (Supplementary Fig.~\ref{fig:ablation}B). HOLMES instead organizes stimuli sharing a shape along a common higher-level branch while representing more specific variation in descendant leaves. When a novel color is encountered, the hierarchical prior biases the stimulus toward an existing leaf within the appropriate shape-consistent subtree, allowing its prediction to draw on outcome statistics already stored at that reused leaf, even though the exact color has not previously been observed.

The distribution of each model's decision variable relative to the $0.5$ boundary highlights this separation (Fig.~\ref{fig:rule}E). HOLMES's predictions for rewarded- and unrewarded-shape trials form two largely non-overlapping distributions, cleanly separated across the boundary in the direction consistent with the true reward contingency. The flat model's two distributions, by contrast, overlap substantially and sit much closer to the boundary itself. Together, these results indicate that the hierarchical prior supports genuine one-shot forward transfer: HOLMES can generalize a learned reward rule to a stimulus it has never encountered, using only the shared higher-level structure of the tree, whereas the flat model has no comparable route to abstraction and must treat each new color as effectively unfamiliar.

\begin{manuscriptfigure}{fig:rule}{\includegraphics[width=1\linewidth,height=0.67\textheight,keepaspectratio]{NECO-D-26-00058R2-Figure.4.eps}}
\manuscriptcaption{Forward transfer to novel colors under a shape-only reward rule ($n=200$ seeds; the same shape was rewarded in every seed).
    \textbf{(A)} Task structure. Reward depends only on shape and is invariant to color. Each of the two shapes was presented in eight training colors and tested in two held-out colors. The critical measurement was the model's prediction on the first presentation of each held-out shape--color pairing, before feedback on that trial.
    \textbf{(B)} One-shot accuracy on the four held-out shape--color pairings per seed. Each model's predicted probability of reward was thresholded at $p=0.5$, and the resulting binary decision was scored against the actual reward label. Mean accuracy was $0.544 \pm 0.032$ for the flat model and $0.926 \pm 0.020$ for HOLMES (95\% confidence intervals across seeds).
    \textbf{(C)} ROC curves for the same predictions, pooled across 800 held-out trials. The rewarded shape was the positive class, and the score was the model's predicted probability of reward. HOLMES achieved an AUC of $0.984$ (95\% CI $[0.977, 0.990]$), and the flat model achieved an AUC of $0.571$ (95\% CI $[0.532, 0.607]$). Confidence intervals were computed by bootstrap resampling over seeds.
    \textbf{(D)} Example learned cluster structures. A hierarchical representation that groups by shape and treats color as a nested, exchangeable leaf-level feature supports immediate generalization to a novel color of a known shape. A representation organized by color instead treats every novel color as an unfamiliar category, providing no basis for transfer.
    \textbf{(E)} Distributions of predicted reward probabilities relative to the $p=0.5$ decision boundary, shown separately for held-out trials of the rewarded and unrewarded shapes. HOLMES separates the two distributions more clearly around the decision boundary, while the flat model's distributions largely overlap.
    }
\end{manuscriptfigure}

While both models used an identical leaf-level read-out of reward, the prior over assignments allows HOLMES to build multidimensional structure. The flat model represents a latent cause as a single cluster over the full feature vector and, during training, forms clusters tied to specific shape--color conjunctions; a held-out color matches no existing cluster on its color feature, so its prediction regresses toward the prior and the shape contrast collapses to zero. 
HOLMES instead organizes training stimuli so that stimuli sharing a shape
tend to traverse common higher-level branches while color-specific
variation is represented in descendant structure. Through pooled
branch-selection statistics and stickiness, the hierarchical prior biases a
novel-color stimulus toward an existing leaf within the corresponding
shape-consistent subtree, where it can reuse previously accumulated outcome
statistics.
The flat prior affords no comparable route to a shape-level abstraction,
and so cannot generalize before receiving feedback on the new color. This advantage also did not depend on a particular parameter setting (Supplementary
Fig.~\ref{fig:sup_adv}E--H).

\section*{Discussion}

We introduced HOLMES, an online hierarchical latent-cause model for sequential structure learning. The model combines hierarchical nonparametric structure with tractable sequential inference, allowing latent causes to be organized across multiple levels of abstraction during online learning. Across compositional tasks, the hierarchical model matched the predictive performance of flat latent-cause models while learning more compact representations and enabling improved one-shot transfer. In a rule-dependent task, hierarchical inference additionally improved one-shot forward transfer to unseen exemplars. These results demonstrate how hierarchical latent structure can support both efficient representation and flexible generalization in sequential learning. 

Hierarchical Bayesian models, such as 
 the nested Chinese Restaurant Process (nCRP) \citep{griffiths2003hierarchical,blei2010nested}, originally developed for hierarchical topic modeling, can represent rich multi-level structure but typically rely on batch inference. Online latent-cause inference models \citep{gershman_context_2010} perform sequential inference but operate over flat latent spaces. By combining a modified nCRP prior with particle filtering, HOLMES extends online latent-cause inference to hierarchical representations whose occupied branching structure and effective depth are inferred sequentially from experience. Incoming observations can occupy latent paths of different depths, allowing the degree of hierarchical elaboration to emerge during inference rather than being fixed in advance for a particular task architecture.
Unlike hierarchical models that rely on batch inference \citep{griffiths2003hierarchical}, architectures whose number of hierarchical tiers is specified for the task class \citep{wang2011online,Liu2022}, or Bayesian model selection over candidate structures \citep[e.g., the COIN model,][]{Heald2021}, HOLMES performs online inference over both latent assignments and effective hierarchical depth, enabling hierarchical representations to be elaborated dynamically during sequential experience.

Our simulations illustrate two functional consequences of hierarchical inference. In compositional environments, hierarchical structure learning produced substantially more efficient representations while preserving outcome prediction performance. These compressed representations aligned with the latent generative structure of the task and enabled rapid generalization across observations sharing higher-level categories. The flat latent cause model, in contrast, required multiple independent clusters to represent the same structure, limiting their ability to support transfer across related observations. This result highlights how representational organization, rather than predictive accuracy alone, can determine the range of generalization operations supported by a learned model.

% add discussion of the replacement Figure 4 task and why the flat model can struggle even though the abstract solution is flat
Further, the shape--color task provides a direct test of forward transfer. The correct abstract rule can be represented as two shape-defined reward classes, so the task is not impossible for a flat model in principle. The difficulty is inferential: the flat model must discover that abstraction online from full stimulus vectors containing both reward-relevant shape features and reward-irrelevant color features. HOLMES can separate these demands across levels of abstraction, allowing shape-level reward structure to be reused while color-specific variation is represented at lower levels. This result clarifies that the hierarchical advantage arises from the ability to discover and reuse abstractions under online inference, rather than from giving the model an intrinsically more powerful outcome readout.

Prior work has shown that one-shot generalization can arise from richly structured generative priors \citep[e.g.,][]{Lake2015}. The open question, however, is how reusable abstractions that support one-shot generalization can be discovered incrementally from sequential experience without explicit supervision over the latent structure. Our contribution is therefore not one-shot learning per se, but a tractable online framework for inferring hierarchical latent representations that make one-shot transfer possible. Unlike standard one-shot concept-learning benchmarks, the task studied here requires category structure to be inferred from unlabeled sequential experience rather than specified in advance by a richly structured concept prior.

A key refinement of our model is that depth is inferred rather than fixed. We constrain $\alpha$ to decay with tree depth. This can be interpreted as a bounded-rational \citep{Bhui2021} prior over structure: agents cannot ``over-carve'' the world at every scale but must allocate a limited segmentation budget. This constraint is consistent with the idea that abstraction is economical—higher levels of representation are sparse and conservative, while finer distinctions emerge only when prediction errors justify additional complexity. This depth-dependent attenuation implements a soft penalty on hierarchical elaboration, capturing the trade-off between expressive capacity and representational simplicity.
More broadly, the model connects to theories of approximate Bayesian inference under resource constraints \citep{Simon1957,Gershman2015,Bhui2021}. Particle filtering approximates the Bayesian posterior using limited computational resources, introducing a trade-off between accuracy and computational cost: the number of particles determines how well the filter can represent multimodal hierarchical beliefs \citep{Findling2020}.

 %%% discuss Liu

A related line of work has examined how learners represent and reuse structured task knowledge \citep{Franklin2018, Franklin2020, Liu2022}. Franklin and Frank (2020) showed that human learners can flexibly rely on compositional or conjunctive representations depending on task structure. Liu and Frank (2022) extended this computational approach within a hierarchical Bayesian framework, allowing structured task components and their dependencies to support transfer and generalization across environments. This framework represents hierarchical task structure, while the hierarchical architecture appropriate to a given task class is specified as part of the model.

HOLMES addresses a complementary question: whether the occupied branching structure and effective depth of a nested latent representation can themselves be inferred online from sequential experience. The two frameworks also differ in their representational commitments. Liu and Frank explicitly represent distinct task components, allowing their structure and dependencies to be learned and reused compositionally \citep{Liu2022}. HOLMES instead learns nested structure over observations as a whole, without explicitly factorizing observations into independently reusable components. Thus, the two approaches address distinct but complementary aspects of latent structure learning. HOLMES focuses on the online construction of hierarchical depth, whereas Liu and Frank build compositional organization and reuse of task structure. A natural extension would combine these capabilities, allowing independently reusable task components to develop their own nested hierarchies while also learning dependencies among those components.

% added limitation section to address reviewer q on non-tree/compositional-structure
Several limitations remain. Our analysis focused on synthetic tasks with discrete binary features and latent structure that can be compactly expressed as a tree. This is a natural and important class of structure, but it is not exhaustive. Some environments are better described by factorial or directed-acyclic graphical structures, in which multiple independent factors jointly determine observations or outcomes~\citep{Kemp2008, Ghahramani1997}. Such settings are not naturally represented by a single nested tree. Liu and Frank (2022)’s framework addresses this limitation by representing distinct task components, such as mappings and rewards, separately and allowing their structure and dependencies to be learned and reused across environments. Thus, the two approaches provide complementary capabilities: HOLMES enables online inference over hierarchical depth, while Liu and Frank provide a richer framework for  compositional structure learning and generalization.

In real-world environments, hierarchical structure may also be ambiguous, partially overlapping, or only weakly identifiable from early experience. Under such conditions, the hierarchical prior can sometimes lead to negative transfer: early commitment to an incorrect abstraction may cause later observations to generalize along the wrong dimension, while excessive fragmentation may prevent useful transfer (Fig.~\ref{fig:neg_transfer}). These failure modes may be useful to examine edge cases in which human inference also breaks down. 

Despite these limitations, our work provides a tractable computational framework for online hierarchical latent-cause inference, offering a foundation for studying rule discovery, one-shot generalization, and structural learning in both biological and artificial learning systems. 

\paragraph{Code availability.} All model and simulation code is available at \url{https://github.com/Iigaya-Lab/HOLMES-2026}.

All simulations were implemented in Python using NumPy for numerical operations, Scipy and Sklearn for statistics, and figures were generated using Matplotlib and Seaborn. 

\section*{Acknowledgments}

The authors would like to thank Kim Stachenfeld, Adithya Gungi, Deniz Yagmur Urey, Pradyumna Sepulveda, and Sashank Pisupati. Research reported in this publication was supported by the Brain \& Behavior Research Foundation Young Investigator Grant (KI), and the National Institute of Neurological Disorders and Stroke (T32NS064929; IA).

\clearpage
% APA-formatted bibliography compiled from the cited records in crps.bib.
% APA-formatted references from the 36 cited records in crps.bib.

\clearpage

\setcounter{figure}{0}
\renewcommand{\thefigure}{S\arabic{figure}}
\setcounter{page}{1}
\renewcommand{\thepage}{S-\arabic{page}}
\renewcommand\thesection{S\arabic{section}}
\setcounter{subsection}{0}
\setcounter{section}{0}

\section*{Supplemental Materials}
\subsection*{Extended Methods}

\subsubsection*{Particle Filter}
Exact posterior inference over latent assignments is intractable in both the flat and hierarchical models. We therefore approximate the posterior using a particle filter with $P = 200$ particles, which we found to provide stable estimates without unnecessary computational cost.

A convergence check across
$n_{\text{particles}} \in \{25, 50, 100, 200, 400, 800\}$,
holding the
trial sequence fixed across 20 repeated inference runs per particle
count with independent random seeds, showed the mean estimate at
$n_{\text{particles}}=200$ differed from the estimate obtained with
$4\times$ as many particles ($n_{\text{particles}}=800$) by no more than
one standard error of the mean in both models (HOLMES: $\Delta = 0.005$,
$\text{SEM} = 0.005$; flat: $\Delta = 0.002$, $\text{SEM} = 0.002$),
indicating that the estimate had stabilized well within the range
tested.

Each particle $p \in \{1, \dots, P\}$ maintains a cluster assignment $Z_t^{(p)}$ sampled from the CRP prior, together with feature count statistics $\{n_{f,k}^{(p)}, b_{f,k}^{(p)}\}$ for each cluster $k$ discovered by that particle. Each particle's weight $w_t^{(p)}$ represents how well that hypothesis explains the observed data. Since particles are sampled directly from the CRP prior rather than from a separate proposal distribution, the weight reduces to the data likelihood under that particle's cluster assignment:
\begin{equation}
    w_t^{(p)} = P(\mathbf{d}_t \mid Z_t^{(p)}).
\end{equation}
Weights are normalized so that $\sum_p w_t^{(p)} = 1$, forming a discrete approximation to the posterior. Particles are resampled at each time step with probability proportional to their weights, concentrating computational resources on high-likelihood hypotheses and pruning implausible ones, thereby preventing particle degeneracy~\citep{Kantas2009, 1521264}. Hypotheses that predict data well proliferate, while poor hypotheses are eliminated. The model's predictions are computed as posterior expectations over the particle ensemble, naturally reflecting uncertainty when particles disagree about cluster assignments: when particles are distributed across different cluster assignments, the prediction is a mixture of different cluster statistics.

For HOLMES, each particle instead maintains a sampled path
$\mathbf{c}_t^{(p)}$ through the hierarchical prior, together with
leaf-level feature statistics. The terminal node of the sampled path is
mapped to the leaf cause used for likelihood evaluation and prediction.

\begin{manuscriptfigure}{fig:crp-sup}{%
    \includegraphics[width=1\linewidth,height=0.67\textheight,keepaspectratio]{NECO-D-26-00058R2-Figure.S1.eps}}
    \manuscriptcaption{Flat and Hierarchical model sequential learning.
(A) One-layer (``flat'') model. Standard CRP-based models maintain a single partition over observations, incrementally assigning each observation to an existing cluster or creating a new one. Trial 1: The first observation creates a new group. Trial 2: The second observation (dashed) creates another new group. Trial 3: The third observation is assigned to the existing compatible group. Trial n: After many trials, the model has discovered multiple groups at a single level of abstraction.
(B) Online hierarchical latent structure learning. Our model generalizes the flat formulation by organizing latent structure across multiple levels of abstraction \textit{online}. The hierarchical extension builds tree structures across multiple levels. 
Trial 1: The first observation creates new nodes at multiple levels (each marked 'new'). Trial 2: The second observation creates a new branch, forming a sibling relationship with the first observation at higher levels while differing at lower levels, and deepening the hierarchy. Trial 3: The third observation reuses existing structure at higher levels but creates new structure at the observation level. Trial n: The model has discovered a multi-level tree with shared structure at higher levels. }
\end{manuscriptfigure}

\paragraph{Depth-decay interpretation.}
The exponential decay schedule in Eq.~\ref{eq:depth_decay} has a natural bounded-rationality interpretation: it implements a fixed segmentation budget that must be allocated across levels of the hierarchy. Higher $\alpha$ exhausts this budget quickly, producing shallow trees; lower $\alpha$ preserves budget for deeper levels, encouraging finer-grained distinctions when evidence warrants them. This adaptive depth control emerges automatically from the concentration schedule without requiring an explicit depth limit. The combination of depth decay and stochastic stopping (Eq.~\ref{eq:stopping}) provides two complementary mechanisms for controlling tree complexity: depth decay exponentially reduces the prior probability of branching at deeper levels, while stochastic stopping allows the model to terminate paths early even when branches exist. Together, they implement a flexible prior over hierarchical depth that adapts to environmental structure.

% \paragraph{Likelihood computation.}
% Likelihoods are evaluated only at the leaf nodes of the inferred hierarchy. Each leaf maintains Beta-Bernoulli feature statistics $\{n_{f,k}^{(p)}, b_{f,k}^{(p)}\}$, initialized with pseudocount $\Omega$, exactly as in the flat model. After observing feedback, feature counts for the active leaf are updated incrementally. For numerical stability, all likelihood computations are performed in log-space.

\paragraph{Likelihood computation.}
Although likelihoods are evaluated only at the leaf nodes, upper levels shape inference through the prior: at each level $\ell$, the nested CRP assigns observations to branches with probability proportional to their occupancy, so observations routed through the same ancestor have elevated prior probability of sharing a descendant leaf and therefore pool their sufficient statistics. Upper levels thus determine which leaves are reachable and with what probability, without contributing directly to the likelihood computation.

\paragraph{Canonical node reuse.}
When different particles discover identical substructures during inference, they share the same global node identity rather than maintaining redundant local copies. This canonical node reuse~\citep{blei2010nested} supports hierarchical abstraction by ensuring that the inferred tree represents shared latent structure across the particle ensemble, rather than a collection of particle-specific local trees. Global node identities are assigned at the time a new branch is first created and are preserved across all particles that subsequently visit the same branch.

\paragraph{Stickiness}
To encourage temporal persistence, we add a stickiness bonus \citep{Fox2011} that increases the probability of reusing the branch selected on the previous trial. Let $k^* = c_{t-1}^{(p,\ell)}$ denote the branch chosen at level $\ell$ on the previous trial (after resampling), and let $n_{k}^{(p,\ell)}$ be the pooled branch-occupancy count defined in Eq.~\ref{eq:ncrp_existingnew}. Before normalizing, the existing-branch count for the previously taken branch is scaled multiplicatively:
\begin{equation}
    \tilde{n}_{k}^{(p,\ell)} = n_{k}^{(p,\ell)} \cdot (1 + \Omega)^{\mathbb{1}[\,k = k^*\,]},
    \label{eq:stickiness}
\end{equation}
so the previously taken branch is up-weighted by a factor $(1+\Omega)$, while non-matching existing branches and the new-branch term $\alpha_\ell$ are left unchanged; probabilities are then renormalized to sum to one. $\Omega$ thus serves a dual role as the Beta--Bernoulli pseudocount and the stickiness strength.

\subsubsection*{Simulations}

We instantiate the hierarchical latent-cause model parameterized by the concentration parameter (and depth limiter) $\alpha$ and the observation pseudocount (and stickiness) $\Omega$ as described above. We used 200 particles for all simulations.

Our operating point of 200 particles further sits within the range commonly used in cognitively motivated latent-cause and categorization models: prior work has used as few as 5 particles \citep{Harhen2021}, 100 particles \citep{Sanborn2006, Sanborn2010} to describe human and animal behavior, while other latent-cause models have used substantially larger particle counts (e.g., 3,000 for full inference, but 100 in figures; \citealp{gershman_context_2010}) explicitly to approximate exact inference. Our choice reflects a point at which additional particles no longer meaningfully change model predictions, consistent with the view that bounded sample-based inference is itself a plausible feature of cognitive computation rather than a shortcoming \citep{Findling2020}.

For HOLMES modeling, we set a maximum tree depth of 20 levels and a maximum of 20 children per node to limit runaway growth while retaining substantial representational capacity.

\subsubsection*{Tasks}
To systematically assess the representational efficiency of hierarchical versus flat structure, we generated synthetic tasks with controlled hierarchical ground truth and varying complexity.

\paragraph{HOLMES reduces to flat latent-cause behavior when hierarchy is not warranted}

\begin{figure}
    \centering
    \includegraphics[width=1\linewidth,height=0.54\textheight,keepaspectratio]{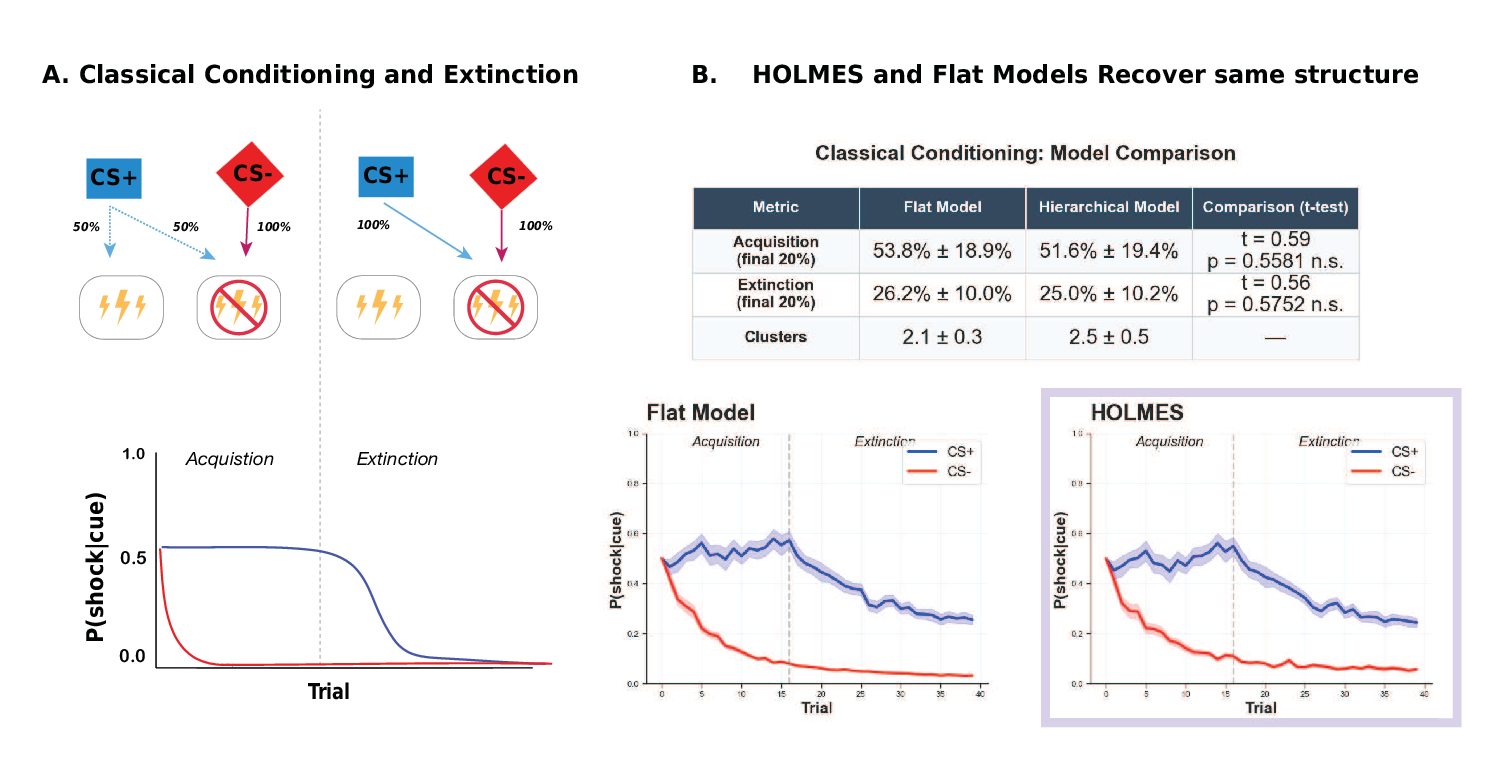}
    \caption{HOLMES recovers standard latent-cause behavior when no hierarchical structure is present.
    \textbf{(A)} Classical conditioning task structure. During acquisition, CS+ is followed by shock on 50\% of trials and by no shock on the remaining 50\%, while CS- is never followed by shock. During extinction, shock is withheld entirely, so both cues are followed by no shock on 100\% of trials. The idealized associative-strength trace (bottom) shows the qualitative prediction: $P(\text{shock}\mid\text{CS-})$ falls rapidly during acquisition as the deterministic non-reinforcement contingency is learned, while $P(\text{shock}\mid\text{CS+})$ remains near the true reinforcement rate ($\sim$0.5) throughout acquisition and decays toward zero once shock is removed during extinction.
    \textbf{(B)} Model comparison (n = 50 seeds). Top: asymptotic predictions (final 20\% of trials) did not differ between the flat and hierarchical models during acquisition, nor in the number of clusters used, both remaining close to the two true latent groups (CS+, CS-). Bottom: example learning traces of $P(\text{shock}\mid\text{cue})$ across trials for the flat model (left) and HOLMES (right); dashed vertical line marks the acquisition--extinction transition. Both models track statistically indistinguishable associative trajectories for CS+ (blue) and CS- (red).}
    \label{fig:cc}
\end{figure}

The main text results show that HOLMES exploits hierarchical structure when it is present. To test whether the hierarchical prior imposes a cost when there is no true structure, we tested a standard classical conditioning and extinction task (Fig.~\ref{fig:cc}A) --- the original domain of the latent-cause model \citep{gershman_context_2010} and a setting with no nested latent structure to discover.

HOLMES and the flat model produced statistically indistinguishable asymptotic predictions during both acquisition and extinction, and used a similar, near-minimal number of clusters (Fig.~\ref{fig:cc}B). Because stochastic stopping allows HOLMES's paths to terminate at the first level, the model is not forced to elaborate depth that the data do not support: in this task, its inferred trees remain effectively shallow and its predictions collapse onto those of the flat CRP. This confirms that the hierarchical prior does not distort inference in flat environments, and that HOLMES's advantage in the earlier tasks reflects genuine exploitation of hierarchical structure rather than a general bias toward more complex representations.

\paragraph{Scalable Task structure.}
We constructed a scalable synthetic categorization task in which the number of hierarchical levels $L$ was varied parametrically from 2 to 5. Each level introduced an additional binary latent variable, producing task environments of increasing structural complexity. 

The observation feature space consisted of three types of binary features. First, \textit{observation-level features} (4 dimensions): a one-hot encoding representing the lowest level of the hierarchy, with the feature at index $o \in \{0,1,2,3\}$ set to 1. Second, \textit{first latent features} (2 dimensions, present for $L \geq 2$): both dimensions were assigned the same binary value $\ell_1 \in \{0,1\}$, encoding the first latent category level. Third, \textit{higher latent features} (1 dimension per level above 2, present for $L \geq 3$): each additional hierarchical level contributed one binary feature encoding that level's latent value. 

For a task with $L$ levels, the total number of observable features was $F = \max(0, L-2) + 2 + 4$, giving $F = L + 4$ for $L \geq 2$. The outcome label was appended as a final feature, yielding a feature matrix with dimensions ${(F+1) \times T}$, where $T$ is the total number of trials.

\paragraph{Context enumeration.}
All possible contexts were enumerated recursively. At the base level ($L = 1$), four contexts corresponded to the four observation-level values. At each additional level, the context set doubled by pairing all existing sub-contexts with each of two binary values for the new latent level. This yielded $4 \times 2^{L-1}$ contexts for a task of $L$ levels:

\begin{table}[h]
\centering
\doublespacing
\begin{tabular}{@{}>{\centering\arraybackslash}p{0.27\linewidth}
  >{\centering\arraybackslash}p{0.27\linewidth}
  >{\centering\arraybackslash}p{0.35\linewidth}@{}}
\hline
\textbf{Hierarchical Levels ($L$)} & \textbf{Number of Contexts} & \textbf{Trials per Level (10 per context)} \\
\hline
2 & 8  & 80  \\
3 & 16 & 160 \\
4 & 32 & 320 \\
5 & 64 & 640 \\
\hline
\end{tabular}
\caption{Task complexity scaling with number of hierarchical levels.}
\label{tab:task_complexity}
\end{table}

\paragraph{Outcome rule.}
Outcomes were binary ($y \in \{0,1\}$) and were determined by a conjunctive rule defined over the two highest latent levels of the hierarchy. For $L = 2$, the outcome was $y = 1$ if the first latent level value (level 2) $= 0$, and $y = 0$ otherwise. For $L \geq 3$, the outcome was $y = 1$ if and only if both the top latent level value (level $L$) $= 0$ \textit{and} the second-highest latent level value (level $L-1$) $= 0$; otherwise $y = 0$. 

This conjunction rule ensured that no single observation-level feature was sufficient to predict the outcome: models were required to identify and represent the relevant higher-order latent structure to achieve high outcome prediction accuracy.

\paragraph{Observation generation and noise.}
For each context, a prototype feature vector was constructed according to the encoding rules above. On each trial, a copy of the prototype was presented with a small probability of perceptual noise: with probability 0.02, one randomly selected observation-level feature dimension was bit-flipped ($0 \rightarrow 1$ or $1 \rightarrow 0$). All trials across all contexts were shuffled uniformly at random prior to presentation using a seeded random number generator, ensuring independent trial orderings across seeds.

Both the flat and hierarchical models received the full feature matrix as input and were evaluated on the same task instance per seed.

% \begin{figure}
%     \centering
%     \includegraphics[width=0.5\linewidth]{supp_binary.png}
%     \caption{Example feature vector encoding for 2-level task structure.}
%     \label{fig:binary}
% \end{figure}
\paragraph{Outcome prediction accuracy.}
The primary learning metric was outcome prediction accuracy, computed as the proportion of trials on which the model's binarized outcome estimate matched the true outcome:
\begin{equation}
    \text{Acc}_\text{outcome} = \frac{1}{T} \sum_{t=1}^{T} \mathbf{1}\left[[\hat{r}_t > 0.5] = y_t\right]
\end{equation}
where $\hat{r}_t$ is the model's posterior mean outcome estimate at trial $t$ and $y_t \in \{0,1\}$ is the true outcome. This was computed separately for the flat model ($\hat{r}_t^\text{flat}$) and the hierarchical model ($\hat{r}_t^\text{hier}$).

To characterize the representational efficiency of each model, we computed two key metrics relating the model's learned cluster structure to the ground-truth categorical labels at each hierarchical level. For each trial, the cluster assignment was determined by majority vote across particles. Trials with no valid assignment (cluster id $= -1$) were excluded, with analyses requiring at least 10\% valid coverage and a minimum of 5 valid trials.

\begin{itemize}
    \item \textbf{Number of clusters} ($K$): the total number of distinct cluster identities present in the valid majority-voted trial assignments.
    
    \item \textbf{Average within-label entropy}: for each ground-truth label $m$, we computed the Shannon entropy of the cluster-assignment distribution over valid trials with that label,
    \[
        H_m = -\sum_k p_{mk}\log p_{mk},
    \]
    where $p_{mk}$ is the proportion of label-$m$ trials assigned to cluster $k$. We reported the frequency-weighted average across labels,
    \[
        \bar{H} = \sum_m w_m H_m,
        \qquad
        w_m = \frac{N_m}{\sum_{m'} N_{m'}},
    \]
    where $N_m$ is the number of valid trials with label $m$. Lower values indicate that trials sharing the same ground-truth label are assigned more consistently to a smaller set of clusters.
\end{itemize}

For the flat model, representational efficiency metrics were computed directly from the particle filter's cluster assignments. For the hierarchical model, representational efficiency metrics were computed using the same tree level that was selected for transfer, ensuring consistency between the generalization and representational efficiency assessments.

\paragraph{Parameter selection.}

Backward transfer scaling task analyses were performed across 200 randomly sampled parameter combinations ($\alpha \in [0.1, 3.0]$, $\Omega \in [0.1, 3.0]$), with each combination evaluated over 6 independent random seeds. For each parameter combination, we computed mean performance across its 6 seeds. Reported statistics (confidence intervals, effect sizes) were computed across these 200 parameter combinations, capturing variability across the parameter space rather than seed-to-seed variability alone. The asymptotic outcome prediction accuracy was computed from the final 70\% of trials. Error bars represent 95\% confidence intervals unless otherwise noted. The forward transfer were also across 200 randomly sampled parameters with 6 seeds each in the range ($\alpha \in [0.1, 1.0]$, $\Omega \in [0.1, 0.5]$).

\paragraph{Compression efficiency is robust}
Hierarchical models showed superior compression efficiency (lower entropy) across 100\% of tested parameter combinations (mean entropy advantage: $-0.677$, range: $[-0.990, -0.155]$; Figure ~\ref{fig:sup_adv}, panels A-B). The negative values indicate that hierarchical models consistently achieved lower entropy than flat models, demonstrating more efficient category-to-cluster mappings regardless of parameter settings. This universal advantage indicates that hierarchical compression reflects an architectural property rather than parameter-specific tuning.

\paragraph{One-shot transfer generalization.}
To assess whether each model had learned a representation that supported generalization, we evaluated one-shot categorization transfer after task completion. For each hierarchical level $\ell \in \{2, \ldots, L\}$, the true factor at that level partitioned trials into two categories (factor value 0 or 1). A single labeled exemplar was selected as the first trial in the sequence for which the factor value was 0. The model's cluster assignment at that trial was taken as the anchor cluster. All remaining trials were then classified as belonging to the same category as the anchor (predicted category 0) if assigned to the same cluster, or to a different category (predicted category 1) otherwise.

Transfer performance was quantified using three metrics that characterize different aspects of generalization. \textbf{Recall} measures the proportion of true same-category trials correctly identified as such:
\begin{equation}
    \text{Recall}^{(\ell)} = \frac{\text{TP}}{\text{TP} + \text{FN}} = \frac{1}{|\mathcal{T}_0|} \sum_{t \in \mathcal{T}_0} \mathbf{1}\left[\hat{c}_t = \hat{c}_\text{anchor}\right]
\end{equation}
where $\mathcal{T}_0 = \{t : z_t^{(\ell)} = 0\}$ is the set of trials belonging to true category 0 at level $\ell$, $\hat{c}_t$ is the majority-voted cluster assignment of trial $t$ across particles, $\hat{c}_\text{anchor}$ is the majority-voted assignment of the labeled exemplar trial, TP denotes true positives (correct same-category identifications), and FN denotes false negatives (same-category trials misclassified as different). \textbf{Precision} quantifies the specificity of same-category predictions: Precision = TP/(TP + FP), where FP denotes false positives (different-category trials misclassified as same). \textbf{F1 score} is the harmonic mean of precision and recall, balancing both aspects: F1 = 2 $\times$ (Precision $\times$ Recall)/(Precision + Recall).

We report recall as the primary metric because it directly quantifies the model's generalization ability—the capacity to recognize same-category instances across contexts—independent of overgeneralization tendencies. Category membership is balanced in our task design, making recall directly interpretable. Critically, recall is highly correlated with F1 score across all parameter combinations (Table~\ref{tab:transfer_metrics} note), confirming that conclusions are robust to metric choice. Precision and F1 provide complementary information about the precision-recall tradeoff.

For the hierarchical model, cluster assignments were extracted from the tree path at a target depth corresponding to the level being evaluated ($\text{depth} = \ell - 1$). A fallback procedure was applied: if fewer than 10\% of trials had valid (non-negative) node assignments at the target depth, the model fell back to the next shallower level, continuing until a sufficient level was found or no valid level remained.

Table~\ref{tab:transfer_metrics} presents transfer performance across all metrics and task complexities.

\begin{table}[htbp]
\centering
\doublespacing
\caption{Transfer generalization metrics by model and task complexity}
\label{tab:transfer_metrics}
\begin{tabular}{clccc}
\toprule
\textbf{Level} & \textbf{Model} & \textbf{Recall} & \textbf{Precision} & \textbf{F1} \\
\midrule
\multirow{2}{*}{2} 
&\textbf{Flat}& 0.897 (0.147) & 0.995 (0.025) & 0.918 (0.117) \\
& Hierarchical & 0.893 (0.128) & 0.656 (0.097) & 0.707 (0.106) \\
\midrule
\multirow{2}{*}{3} 
& Flat         & 0.445 (0.064) & 0.661 (0.132) & 0.507 (0.064) \\
& \textbf{Hierarchical}& 0.655 (0.137) & 0.579 (0.091) & 0.541 (0.080) \\
\midrule
\multirow{2}{*}{4} 
& Flat         & 0.329 (0.098) & 0.517 (0.076) & 0.368 (0.080) \\
& \textbf{Hierarchical}& 0.576 (0.186) & 0.601 (0.077) & 0.468 (0.114) \\
\midrule
\multirow{2}{*}{5} 
& Flat         & 0.355 (0.085) & 0.533 (0.077) & 0.397 (0.068) \\
& \textbf{Hierarchical}& 0.621 (0.186) & 0.590 (0.076) & 0.507 (0.110) \\
\bottomrule
\end{tabular}
\begin{minipage}{\linewidth}
\begin{tablenotes}
\item  Mean (SD) across 200 parameter combinations (6 seeds each). At 3--5 levels, hierarchical models achieve substantially higher recall and F1 scores, indicating superior generalization with balanced precision-recall tradeoff. Recall and F1 are highly correlated across parameter combinations (3-level: $r = 0.85$; 4-level: $r = 0.95$; 5-level: $r = 0.95$).
\end{tablenotes}
\end{minipage}
\end{table}

\begin{figure}
    \centering
    \includegraphics[width=1\linewidth,height=0.66\textheight,keepaspectratio]{NECO-D-26-00058R2-Figure.S3.eps}
    \caption{Transfer Advantage Parameters. Parameter sweep for both parameters between 0.1-3.0, with 200 randomly sampled parameter values in this range. Parameter heatmap shown with scores on (A) task accuracy as well as average transfer advantage. Stars show the parameter regime that conferred the greatest accuracy or recall advantage, respectively. (B) Transfer accuracy on all levels tested. Brighter colors show higher accuracy. 
    (C) Transfer performance is negatively correlated with task accuracy performance for both models across all levels. This might be because transfer performance encourages generalization which may inhibit early learning in the task. }
    \label{fig:sweep}
\end{figure}

\paragraph{Transfer advantage holds across most parameter space}
Hierarchical models achieved superior backward transfer performance across 94\% of parameter combinations (Supplementary Figure ~\ref{fig:sup_adv}, panels C-D). These results confirm that the hierarchical transfer advantage is robust across nearly all reasonable parameter settings.

\begin{manuscriptfigure}{fig:sup_adv}{%
    \includegraphics[width=0.7\linewidth,height=0.9\textheight,keepaspectratio]{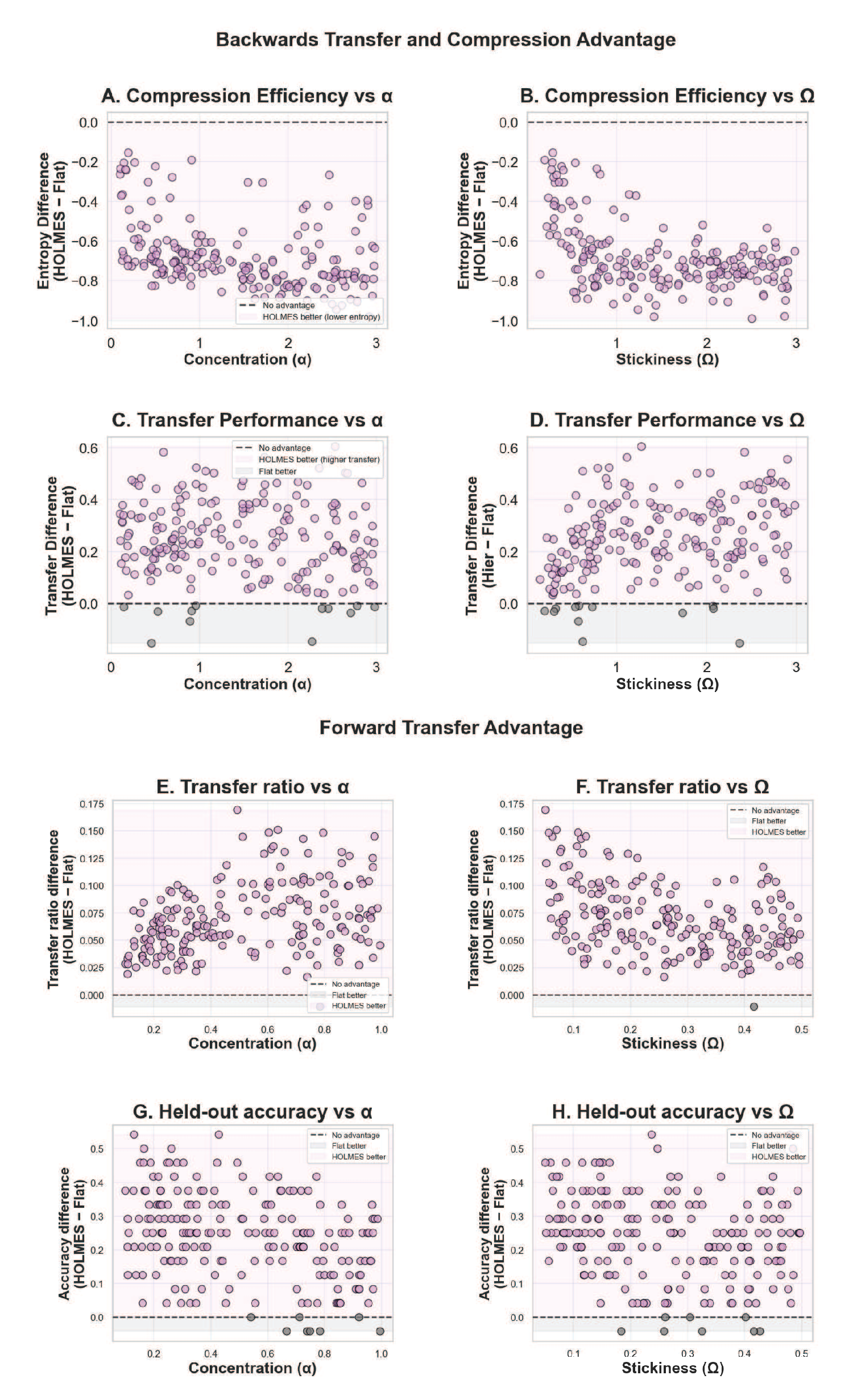}}
    \manuscriptcaption{Parameter sweeps over the concentration parameter $\alpha$ and the stickiness parameter $\Omega$, comparing hierarchical and flat models across $n=200$ parameter combinations per panel.
    \textbf{(A)} Compression efficiency (entropy difference, hierarchical $-$ flat) as a function of $\alpha$. Negative values indicate that the hierarchical model achieves lower entropy (better compression). The hierarchical model showed superior compression in 100\% of tested combinations.
    \textbf{(B)} Compression efficiency as a function of $\Omega$. The hierarchical advantage was universal across all tested $\Omega$ values.
    \textbf{(C)} Transfer performance (transfer difference, hierarchical $-$ flat) as a function of $\alpha$. Positive values indicate that the hierarchical model achieves higher transfer accuracy. The hierarchical model outperformed the flat model in 94\% of parameter combinations.
    \textbf{(D)} Transfer performance as a function of $\Omega$. The hierarchical advantage increased with higher prior concentration and stickiness, reflecting the benefit of stable hierarchical structure.
    \textbf{(E)} Forward-transfer ratio as a function of $\alpha$ (99.5\% of combinations favored the hierarchical model).
    \textbf{(F)} Forward-transfer ratio as a function of $\Omega$.
    \textbf{(G)} Forward-transfer generalization accuracy as a function of $\alpha$.
    \textbf{(H)} Forward-transfer generalization accuracy as a function of $\Omega$ (96\% of combinations favored the hierarchical model).}
\end{manuscriptfigure}

% fixed the rule task

\paragraph{Shape--color generalization task.}
The shape--color task tested whether a learned rule could transfer to novel stimuli. The task contained two shapes and 10 colors. Reward depended only on shape: one shape was rewarded and the other was unrewarded, independent of color. The rewarded shape was randomly selected for each seed and performance contrasts were sign-corrected accordingly, so that results are not an artifact of an arbitrary fixed assignment of reward to a particular shape identity.

The non-outcome feature vector consisted of one-hot copies of shape and color. Shape was encoded with 6 copies of a two-dimensional one-hot code, yielding 12 shape features. Color was encoded with 1 copy of a 10-dimensional one-hot code, yielding 10 color features. No nuisance features were included. The outcome was appended as the final feature, yielding a 23-dimensional feature vector per trial (12 shape + 10 color + 1 outcome). Both models received the identical feature matrix, the same feedback schedule, and the same online prediction target.

The 10 colors were divided into 8 training colors and 2 held-out test colors. During training, every pairing of the two shapes with each of the 8 training colors was presented 10 times (160 training trials total), shuffled within each block. At test, each of the 2 held-out colors was paired with each shape and presented 3 times, in shuffled order. The critical transfer trials were the \emph{first} presentations of each held-out color--shape pairing, before any feedback had been received for that color --- 4 such trials per seed (2 shapes $\times$ 2 held-out colors). Length-normalized likelihoods were used in this task to avoid making the posterior depend on the arbitrary number of binary feature copies used to encode shape or color; the same normalization was applied to both models.

\paragraph{Evaluation metrics.}
The primary metrics were first-exposure held-out outcome prediction accuracy and area under the ROC curve (AUC). Accuracy thresholds each model's posterior predictive probability of reward at $p = 0.5$ and scores the resulting binary decision against the true (sign-corrected) reward contingency. AUC instead asks a threshold-free question: given one held-out trial from the rewarded shape and one from the other shape, what is the probability that the model assigns a higher predicted reward probability to the rewarded-shape trial? Equivalently, AUC $= P(\text{score}_{\text{rewarded shape}} > \text{score}_{\text{other shape}})$, where score is each model's predicted $p(\text{reward})$ on that trial. This is useful here because it separates two possible failure modes: a model can fail to threshold correctly at $p=0.5$ while still ranking rewarded- and unrewarded-shape trials correctly on average, and AUC will detect that residual signal where thresholded accuracy will not. We therefore report pooled AUC, computed by pooling all first-exposure held-out trials across seeds (with items relabeled so that the rewarded shape is always the positive class), together with a 95\% confidence interval obtained by bootstrap resampling over seeds (2000 resamples). 

These results were also stable across sampled parameter regimes (Supplementary Figure ~\ref{fig:sup_adv}, panels E-H). 

\paragraph{Failure Modes.}
To distinguish genuine negative transfer from failures of structural inference, we constructed a matched reversal probe based on the shape--color forward-transfer task. Models were first trained on the same shape-dependent reward rule used in the main forward-transfer analysis. Test stimuli combined the familiar shapes with a held-out color. In the congruent condition, the shape--reward contingency remained identical to training; in the reversed condition, the contingency was inverted. The stimulus presented on the first test trial was matched across conditions, and the prediction was generated before any feedback from the novel context was incorporated. Consequently, any difference in whether this initial prediction was adaptive or maladaptive reflects the previously learned inductive bias rather than information acquired during test.

We quantified the direction and strength of each prediction as its signed distance from the $p=0.5$ decision boundary, oriented so that positive values favored the currently correct outcome and negative values favored the incorrect outcome. We then continued to provide trial-by-trial feedback in the reversed condition to measure adaptation. This design therefore tests two signatures of negative transfer: whether a bias that improves one-shot prediction when the learned rule remains valid produces systematic error when that rule is reversed, and whether the model subsequently revises that prediction as contradictory evidence accumulates.

\begin{manuscriptfigure}{fig:neg_transfer}{%
    \includegraphics[width=0.9\linewidth,height=0.67\textheight,keepaspectratio]{NECO-D-26-00058R2-Figure.S5.eps}}
\manuscriptcaption{\textbf{Learned inductive biases produce both positive and negative transfer.}
(A) Negative-transfer task structure. Models were first trained on a rule-structured shape--color task in which reward depended on shape while color varied (main text Fig.4). At test, familiar shapes were presented in a novel color under one of two matched contingencies. In the \emph{congruent} condition, the training shape--reward rule remained valid; in the \emph{incongruent/reversed} condition, the shape--reward mapping was flipped. The first prediction in the novel context was made before any test feedback, so differences in performance reflect transfer of the previously learned bias rather than learning within the new context.
(B) First-trial transfer. Signed prediction margin is positive when the prediction favors the currently correct outcome and negative when it favors the incorrect outcome. In the congruent condition, the flat model showed a mean signed margin of $0.042$ (95\% CI: $[0.020,\,0.065]$), whereas HOLMES showed a larger positive margin (95\% CI: $[0.098,\,0.172]$). When the contingency was reversed, the same learned biases became maladaptive: the flat model showed a higher mean margin than HOLMES. Thus, the stronger bias that benefits HOLMES when the learned rule remains valid produces correspondingly stronger negative transfer when that rule is violated. Error bars show 95\% confidence intervals across 100 simulation seeds.
(C) Post-reversal adaptation. With continued feedback under the reversed contingency, HOLMES progressively revised its predictions, crossed the decision boundary, and ultimately favored the new correct rule. Over the final 10 test rounds, the HOLMES signed prediction margin was well above the decision boundary. Shaded regions show 95\% confidence intervals across 100 simulation seeds. Together, these results show that the same hierarchical inductive bias that supports rapid positive transfer can produce genuine negative transfer when the learned regularity is violated, followed by correction as contradictory evidence accumulates.
}
\end{manuscriptfigure}

\paragraph{Ablations.}
To assess the necessity of each modeling choice, we ran ablation analyses in both tasks described above: the compositional (backward-transfer) task and the shape--color (forward-transfer) task. In each task, we compared five variants holding all other parameters fixed at the values used in the main analyses: the flat model; the flat model with stickiness added; HOLMES with stopping removed; HOLMES with stickiness removed; and the full HOLMES model. This design isolates the independent contribution of hierarchical structure itself, of stopping, and of stickiness, and lets us ask whether their contributions are consistent across the two tasks or task-dependent.

\begin{manuscriptfigure}{fig:ablation}{%
    \includegraphics[width=1\linewidth,height=0.70\textheight,keepaspectratio]{NECO-D-26-00058R2-Figure.S6.eps}}
    \manuscriptcaption{Ablations reveal that stopping and stickiness contribute differently across the two tasks, and that stickiness's effect is not consistently beneficial.
    \textbf{(A)} Compositional (backward-transfer) task. One-shot transfer recall for five model variants---Flat, Flat + stickiness, HOLMES without stopping (forced to fixed maximum depth), HOLMES without stickiness, and the full model---as a function of task complexity (2--5 hierarchical levels; shaded 95\% CI across 25 sampled parameter cells). All HOLMES variants outperform both the flat models.
    \textbf{(B)} Shape--color (forward-transfer) task. First-heldout-trial prediction accuracy ($n=10$ seeds per variant) for the same five variants. Stickiness reduces accuracy in this task. Together, these results indicate that hierarchical structure and stopping generalize their benefit across both backward and forward transfer, whereas stickiness's usefulness is task-contingent.}
\end{manuscriptfigure}

%
% ============================================================

% \paragraph{Use of AI-assisted tools}
% The authors used AI-assisted tools (Claude 4.5 Sonnet, ChatGPT-5.2, GitHub Copilot) for code debugging, refactoring, visualization development, and text editing. All AI-generated outputs were manually verified: revised and commented code was tested for correctness against author created output, visualizations against the raw plotting of simulations, and all text was fact-checked to ensure no accuracy was compromised in edits. All research design, analysis, interpretation, and conclusions represent the authors' original work. No AI tools were used to generate or manipulate primary research data.


\begin{thebibliography}{36}

\bibitem[Aldous(1985)]{Aldous1985}
Aldous, D. J. (1985).
Exchangeability and related topics.
In \emph{École d'été de probabilités de Saint-Flour XIII --- 1983} (pp. 1--198).
Springer Berlin Heidelberg.
\url{https://doi.org/10.1007/bfb0099421}

\bibitem[Anderson(1991)]{anderson1991adaptive}
Anderson, J. R. (1991).
The adaptive nature of human categorization.
\emph{Psychological Review}, \emph{98}(3), 409--429.
\url{https://doi.org/10.1037/0033-295X.98.3.409}

\bibitem[Barak et~al.(2013)]{Barak2013}
Barak, O., Rigotti, M., \& Fusi, S. (2013).
The sparseness of mixed selectivity neurons controls the generalization--discrimination trade-off.
\emph{The Journal of Neuroscience}, \emph{33}(9), 3844--3856.
\url{https://doi.org/10.1523/jneurosci.2753-12.2013}

\bibitem[Bhui et~al.(2021)]{Bhui2021}
Bhui, R., Lai, L., \& Gershman, S. J. (2021).
Resource-rational decision making.
\emph{Current Opinion in Behavioral Sciences}, \emph{41}, 15--21.
\url{https://doi.org/10.1016/j.cobeha.2021.02.015}

\bibitem[Blei et~al.(2010)]{blei2010nested}
Blei, D. M., Griffiths, T. L., \& Jordan, M. I. (2010).
The nested Chinese restaurant process and Bayesian nonparametric inference of topic hierarchies.
\emph{Journal of the ACM}, \emph{57}(2), Article 7.
\url{https://doi.org/10.1145/1667053.1667056}

\bibitem[Buckland \& Gey(1994)]{precision}
Buckland, M., \& Gey, F. (1994).
The relationship between recall and precision.
\emph{Journal of the American Society for Information Science}, \emph{45}(1), 12--19.
\url{https://doi.org/10.1002/(SICI)1097-4571(199401)45:1<12::AID-ASI2>3.0.CO;2-L}

\bibitem[Douc \& Cappe(2005)]{1521264}
Douc, R., \& Cappe, O. (2005).
Comparison of resampling schemes for particle filtering.
In \emph{ISPA 2005. Proceedings of the 4th international symposium on image and signal processing and analysis, 2005} (pp. 64--69).
\url{https://doi.org/10.1109/ISPA.2005.195385}

\bibitem[Findling et~al.(2021)]{Findling2020}
Findling, C., Chopin, N., \& Koechlin, E. (2021).
Imprecise neural computations as a source of adaptive behaviour in volatile environments.
\emph{Nature Human Behaviour}, \emph{5}(1), 99--112.
\url{https://doi.org/10.1038/s41562-020-00971-z}

\bibitem[Fox et~al.(2011)]{Fox2011}
Fox, E. B., Sudderth, E. B., Jordan, M. I., \& Willsky, A. S. (2011).
A sticky HDP-HMM with application to speaker diarization.
\emph{The Annals of Applied Statistics}, \emph{5}(2A), 1020--1056.
\url{https://doi.org/10.1214/10-aoas395}

\bibitem[Franklin \& Frank(2018)]{Franklin2018}
Franklin, N. T., \& Frank, M. J. (2018).
Compositional clustering in task structure learning.
\emph{PLOS Computational Biology}, \emph{14}(4), Article e1006116.
\url{https://doi.org/10.1371/journal.pcbi.1006116}

\bibitem[Franklin \& Frank(2020)]{Franklin2020}
Franklin, N. T., \& Frank, M. J. (2020).
Generalizing to generalize: Humans flexibly switch between compositional and conjunctive structures during reinforcement learning.
\emph{PLOS Computational Biology}, \emph{16}(4), Article e1007720.
\url{https://doi.org/10.1371/journal.pcbi.1007720}

\bibitem[Gershman \& Blei(2012)]{gershman_tutorial_2012}
Gershman, S. J., \& Blei, D. M. (2012).
A tutorial on Bayesian nonparametric models.
\emph{Journal of Mathematical Psychology}, \emph{56}(1), 1--12.
\url{https://doi.org/10.1016/j.jmp.2011.08.004}

\bibitem[Gershman et~al.(2010)]{gershman_context_2010}
Gershman, S. J., Blei, D. M., \& Niv, Y. (2010).
Context, learning, and extinction.
\emph{Psychological Review}, \emph{117}(1), 197--209.
\url{https://doi.org/10.1037/a0017808}

\bibitem[Gershman et~al.(2015)]{Gershman2015}
Gershman, S. J., Horvitz, E. J., \& Tenenbaum, J. B. (2015).
Computational rationality: A converging paradigm for intelligence in brains, minds, and machines.
\emph{Science}, \emph{349}(6245), 273--278.
\url{https://doi.org/10.1126/science.aac6076}

\bibitem[Gershman \& Niv(2012)]{gershman_exploring_2012}
Gershman, S. J., \& Niv, Y. (2012).
Exploring a latent cause theory of classical conditioning.
\emph{Learning \& Behavior}, \emph{40}(3), 255--268.
\url{https://doi.org/10.3758/s13420-012-0080-8}

\bibitem[Ghahramani \& Jordan(1997)]{Ghahramani1997}
Ghahramani, Z., \& Jordan, M. I. (1997).
Factorial hidden Markov models.
\emph{Machine Learning}, \emph{29}(2-3), 245--273.
\url{https://doi.org/10.1023/a:1007425814087}

\bibitem[Ghahramani et~al.(2010)]{ghahramani2010tree}
Ghahramani, Z., Jordan, M., \& Adams, R. P. (2010).
Tree-structured stick breaking for hierarchical data.
\emph{Advances in Neural Information Processing Systems}, \emph{23}.

\bibitem[Griffiths et~al.(2003)]{griffiths2003hierarchical}
Griffiths, T., Jordan, M., Tenenbaum, J., \& Blei, D. (2003).
Hierarchical topic models and the nested Chinese restaurant process.
\emph{Advances in Neural Information Processing Systems}, \emph{16}.

\bibitem[Harhen et~al.(2021)]{Harhen2021}
Harhen, N. C., Hartley, C. A., \& Bornstein, A. M. (2021).
Model-based foraging using latent-cause inference.
In \emph{Proceedings of the annual meeting of the cognitive science society} (Vol. 43).
\url{https://escholarship.org/uc/item/9c33b182}

\bibitem[Heald et~al.(2021)]{Heald2021}
Heald, J. B., Lengyel, M., \& Wolpert, D. M. (2021).
Contextual inference underlies the learning of sensorimotor repertoires.
\emph{Nature}, \emph{600}(7889), 489--493.
\url{https://doi.org/10.1038/s41586-021-04129-3}

\bibitem[Hjort et~al.(2010)]{Hjort_Holmes_Müller_Walker_2010}
Hjort, N. L., Holmes, C., Müller, P., \& Walker, S. G. (2010).
An invitation to Bayesian nonparametrics.
In N. L. Hjort, C. Holmes, P. Müller, \& S. G. Walker (Eds.), \emph{Bayesian nonparametrics} (pp. 1--21).
Cambridge University Press.
\url{https://doi.org/10.1017/CBO9780511802478.001}

\bibitem[Kantas et~al.(2009)]{Kantas2009}
Kantas, N., Doucet, A., Singh, S. S., \& Maciejowski, J. M. (2009).
An overview of sequential Monte Carlo methods for parameter estimation in general state-space models.
\emph{IFAC Proceedings Volumes}, \emph{42}(10), 774--785.
\url{https://doi.org/10.3182/20090706-3-fr-2004.00129}

\bibitem[Kemp \& Tenenbaum(2008)]{Kemp2008}
Kemp, C., \& Tenenbaum, J. B. (2008).
The discovery of structural form.
\emph{Proceedings of the National Academy of Sciences}, \emph{105}(31), 10687--10692.
\url{https://doi.org/10.1073/pnas.0802631105}

\bibitem[Lake et~al.(2015)]{Lake2015}
Lake, B. M., Salakhutdinov, R., \& Tenenbaum, J. B. (2015).
Human-level concept learning through probabilistic program induction.
\emph{Science}, \emph{350}(6266), 1332--1338.
\url{https://doi.org/10.1126/science.aab3050}

\bibitem[Liu \& Frank(2022)]{Liu2022}
Liu, R. G., \& Frank, M. J. (2022).
Hierarchical clustering optimizes the tradeoff between compositionality and expressivity of task structures for flexible reinforcement learning.
\emph{Artificial Intelligence}, \emph{312}, Article 103770.
\url{https://doi.org/10.1016/j.artint.2022.103770}

\bibitem[Mansinghka et~al.(2006)]{mansinghka2006structured}
Mansinghka, V. K., Kemp, C., Tenenbaum, J. B., \& Griffiths, T. L. (2006).
Structured priors for structure learning.
In \emph{Proceedings of the twenty-second conference on uncertainty in artificial intelligence} (pp. 324--331).

\bibitem[Martino et~al.(2018)]{Martino2018-st}
Martino, L., Elvira, V., \& Camps-Valls, G. (2018).
Group importance sampling for particle filtering and MCMC.
\emph{Digit. Signal Process.}, \emph{82}, 133--151.

\bibitem[O'Doherty et~al.(2021)]{ODoherty2021}
O'Doherty, J. P., Rutishauser, U., \& Iigaya, K. (2021).
The hierarchical construction of value.
\emph{Current Opinion in Behavioral Sciences}, \emph{41}, 71--77.
\url{https://doi.org/10.1016/j.cobeha.2021.03.027}

\bibitem[Sanborn et~al.(2006)]{Sanborn2006}
Sanborn, A. N., Griffiths, T. L., \& Navarro, D. J. (2006).
A more rational model of categorization.
In R. Sun, G. W. Cottrell, \& N. Miyake (Eds.), \emph{Proceedings of the 28th annual conference of the Cognitive Science Society} (pp. 726--731).
Lawrence Erlbaum.

\bibitem[Sanborn et~al.(2010)]{Sanborn2010}
Sanborn, A. N., Griffiths, T. L., \& Navarro, D. J. (2010).
Rational approximations to rational models: Alternative algorithms for category learning.
\emph{Psychological Review}, \emph{117}(4), 1144--1167.
\url{https://doi.org/10.1037/a0020511}

\bibitem[Simon(1957)]{Simon1957}
Simon, H. A. (1957).
\emph{Models of man: Social and rational}.
Wiley.

\bibitem[Speekenbrink(2016)]{speekenbrink_tutorial_2016}
Speekenbrink, M. (2016).
A tutorial on particle filters.
\emph{Journal of Mathematical Psychology}, \emph{73}, 140--152.
\url{https://doi.org/10.1016/j.jmp.2016.05.006}

\bibitem[Teh et~al.(2006)]{Teh2006}
Teh, Y. W., Jordan, M. I., Beal, M. J., \& Blei, D. M. (2006).
Hierarchical Dirichlet processes.
\emph{Journal of the American Statistical Association}, \emph{101}(476), 1566--1581.
\url{https://doi.org/10.1198/016214506000000302}

\bibitem[Tenenbaum \& Griffiths(2001)]{tenenbaum_generalization_2001}
Tenenbaum, J. B., \& Griffiths, T. L. (2001).
Generalization, similarity, and Bayesian inference.
\emph{Behavioral and Brain Sciences}, \emph{24}(4), 629--640.
\url{https://doi.org/10.1017/S0140525X01000061}

\bibitem[Tenenbaum et~al.(2011)]{tenenbaum2011grow}
Tenenbaum, J. B., Kemp, C., Griffiths, T. L., \& Goodman, N. D. (2011).
How to grow a mind: Statistics, structure, and abstraction.
\emph{Science}, \emph{331}(6022), 1279--1285.

\bibitem[Wang et~al.(2011)]{wang2011online}
Wang, C., Paisley, J., \& Blei, D. M. (2011).
Online variational inference for the hierarchical Dirichlet process.
In \emph{Proceedings of the fourteenth international conference on artificial intelligence and statistics} (pp. 752--760).
JMLR Workshop; Conference Proceedings.

\end{thebibliography}
\end{document}